\documentclass[a4paper,fleqn]{cas-dc}

\usepackage[T1]{fontenc}
\usepackage[english]{babel}
\usepackage{csquotes}      
\usepackage{microtype}

\usepackage[switch]{lineno}

\usepackage{amsmath}
\usepackage{amssymb}
\usepackage{graphicx}
\graphicspath{{Figures/}{Images/}}
\usepackage{subcaption}
\usepackage{orcidlink}
\usepackage{booktabs}
\usepackage{multirow}
\usepackage{makecell}
\usepackage{threeparttable}
\usepackage{array}

\usepackage{algorithm}
\usepackage{algpseudocode}

\usepackage[
  backend=biber,
  style=ieee,
  doi=true,
  url=true,
  isbn=false
]{biblatex}
\begin{document}


\shorttitle{Conditional Diffusion Generation of DVL and Inertial Signals}
\shortauthors{M. Levin and I. Klein}

\title[mode=title]{Conditional Diffusion Generation of DVL and Inertial
Signals for Underwater Navigation}

%

\author[1]{Michal Levin~\orcidlink{0009-0008-1913-4945}}
\fnmark[1]


\ead[url]{mlevin28@campus.haifa.ac.il}


\affiliation[1]{organization={University of Haifa},
    addressline={Abba Khoushy Ave 199}, 
    city={Haifa},
    postcode={3498838}, 
    country={Israel}}

\author[1]{Itzik Klein~\orcidlink{0000-0001-7846-0654}}
\ead{kitzik@univ.haifa.ac.il}
\cortext[cor1]{Corresponding author}

\begin{abstract}
Autonomous underwater vehicles (AUVs) navigate without satellite positioning and rely on Doppler velocity log (DVL) and inertial sensors for motion estimation. Deep learning methods for processing these sensors require training data covering the motion the AUV encounters, commonly obtained through costly sea trials. To overcome this limitation, we present a conditional diffusion framework that generates synthetic DVL and inertial signals to replace real recordings in training. The generator employs an Elucidated diffusion model conditioned on three per-channel motion descriptors. Three configurations are evaluated: DVL beams alone, inertial channels alone, and all channels jointly from one model. Across thirteen trajectories of the real dataset, the generated signals followed the requested conditions with Pearson correlations between 0.72 and 0.997, transferred to trajectories withheld from generator training, and did not replicate any training window. A BeamsNet velocity estimator trained on generated signals alone, with no real recording retained, matched the velocity error obtained with the full real training set. Replacing up to all real windows with generated ones caused no measurable loss. Thus, this approach reduces the dependence of learned AUV navigation on sea-trial data, enables training when recordings are scarce, and facilitates developing and evaluating deep learning methods for underwater navigation.
\end{abstract}


\maketitle

\section{Introduction}
Autonomous underwater vehicles (AUVs) carry a wide variety of sensors to support a broad range of applications, including high-resolution near-bottom data collection in underwater acoustic sensor networks \cite{11616708}, geophysical surveying \cite{oehler2024processing}, and inspection of underwater structures \cite{lopez2026enhanced}. Among these, the sensors enabling autonomous navigation are critical, since the vehicle must localize itself without access to global navigation satellite system (GNSS), which is unavailable underwater since signals are strongly attenuated by seawater \cite{kinsey2006survey}. Underwater navigation therefore depends on onboard sensing, which includes inertial, acoustic, and geophysical techniques to estimate the vehicle's position without external infrastructure \cite{paull2014auv}. In the absence of GNSS, the inertial measurment unit (IMU) and the Doppler velocity log (DVL) form a natural pairing for underwater navigation. Fused, the DVL's velocity measurements bound the growth of inertial drift, while the IMU supplies the high-rate navigation state between DVL updates \cite{stutters2008navigation, titterton2004strapdown, zhang2023autonomous}. A wide range of algorithms address this fusion and the DVL processing that supports it, spanning the estimation of vehicle velocity from raw beam measurements \cite{beamsnet}, the handling of degraded or missing measurements \cite{transbeamsnet,topini2023dvldenied}, the integration of the two sensors within a nonlinear estimation framework to produce an accurate navigation \cite{akit, xu2024invariant, mu2021practical}, and even both tasks with a single algorithm \cite{sahoo2026}. Classical approaches rely on model-based estimation, but learned methods have been increasingly applied to these tasks, as they can capture sensor error patterns and nonlinear behavior directly from data \cite{cohen2024survey}. These methods depend on large volumes of sensor recordings for training, yet such data is costly to obtain. Each recording requires a dedicated AUV sea trial, support vessel, crew, and favorable sea conditions. AUV navigation data is therefore scarce, and the trajectories available for training may be unevenly distributed across motion regimes. This motivates generating synthetic navigation sensor signals that can supplement or replace the recorded data for deep learning methods.

Several strategies have been proposed to address the shortage of training data. A primary approach is data augmentation, which applies random transformations such as noise addition, scaling, and temporal warping to expand existing datasets \cite{iwana2021}, \cite{fekson2026enhancement}. Generative adversarial networks (GANs) have also been commonly used to synthesize signals \cite{cgan}, particularly showing remarkable success in image synthesis \cite{creswell2018}. A GAN learns the distribution of real data indirectly, through an adversarial process between two networks: a generator that synthesizes samples from random noise and a discriminator that learns to distinguish them from real data \cite{alzantot2017}. However, GANs are notoriously difficult to train due to architecture/ agnostic challenges such as vanishing gradients, uninformative loss metrics, and instability \cite{arjovsky2017}. Furthermore, they frequently suffer from mode collapse, a phenomenon in which the generator minimizes its loss by producing only a limited, repetitive subset of samples that successfully fool the discriminator, rather than learning the full diversity of the target distribution, particularly when training data is scarce \cite{Alvaro2022}. 

Diffusion models take a different approach. Rather than learning the distribution through a two-network contest, a diffusion model learns it by reversing a gradual noising process which is trained to denoise signals corrupted across a range of noise levels, and generates new samples by iteratively denoising from pure noise \cite{ho2020}. Since generation starts from noise rather than an existing recording, the samples are not tied to any single window. Training on the single, well-defined task of removing noise also avoids the instability of GAN training described above. While diffusion models first gained prominence in image synthesis, they have since been applied to one-dimensional signals, and to time-series signals in particular, for example for the well-established domain of audio generation \cite{diffwave}, biomedical signals generation \cite{diffecg, patchemg} and recently in inertial sensors generation \cite{oppel2025}. For the latter, diffusion models have been introduced in several forms such as denoising raw gyroscope signals ahead of a learned heading estimator \cite{benarie2026}, learning inertial sensor bias \cite{zhou2025}, and generating synthetic inertial data for training a downstream regression network \cite{cohen2025diffusion}.

Conditional diffusion models are an extension that adds control over generation. In addition to generating samples from the full learned distribution, the conditional model takes a conditioning input and shifts the generation to the region consistent with it \cite{ho2022cfg}. The condition  is supplied to the network throughout the denoising process, acting as a design input that specifies the kind of sample to produce. This turns generation into a guided process, steering the samples to be drawn from the conditional distribution rather than the marginal. 

Inspired from the above inertial diffusion models, this work extends that direction to underwater navigation. To this end, we offer generating the The DVL and the IMU measurements with a conditional diffusion model, under a low-order conditioning on the vehicle motion. The conditioning is chosen to fix a sample to a motion regime while leaving the fine detail to the model, so that generation does not collapse to reproducing a single recording.

The generated signals are evaluated on a downstream navigation task, namely velocity estimation.  To this end, we adopt BeamsNet \cite{beamsnet} a learned replacement for the model-based least-squares solution, trained to compensate for the DVL  measurement errors, with demonstrated improvement over it. Two variants are used, and they differ in the sensors they require:
BeamsNetV1 estimates velocity from DVL beam and IMU measurements
together, while BeamsNetV2 estimates it from DVL beam measurements
alone. The DVL and the IMU observe the same vehicle motion, so the two
streams are not independent, and this work generates them in three
Pipelines. Pipeline 1 generates the beam velocities alone and trains
BeamsNetV2. Pipeline 2 generates the IMU measurements alone and pairs
them with real beam velocities to train BeamsNetV1. Pipeline 3 generates
both sensors jointly from a single model, under a shared conditioning
that ties the two streams to the same motion, and trains BeamsNetV1.

The key contributions of this paper are:

\begin{enumerate}
    \item A conditional diffusion approach for generating AUV navigation
    sensor signals, extending diffusion-based generation of inertial
    training data to the underwater setting.

    \item A conditioning design tailored to this type of signal,
    identifying the motion descriptors that make generation controllable
    while preserving diversity, together with an evaluation protocol
    that quantifies condition adherence, transfer to unseen
    trajectories, draw-to-draw diversity, and non-replication of the
    training data.

    \item Joint generation of the DVL and the IMU readings from a single
    conditional model under a shared conditioning, showing that ten
    correlated channels can be produced together.
\end{enumerate}

The three Pipelines are assessed at two levels on the A-KIT dataset \cite{akit}. At
the signal level, four tests establish whether the conditioning input
reaches the output, whether it transfers to trajectories withheld from
generator training, whether repeated draws from a fixed condition
differ, and whether any generated window reproduces a training window.
At the downstream level, four experiments vary the composition of the
downstream training set, from a fully synthetic set to a fixed-size set
in which real windows are progressively replaced.

A velocity estimator trained on generated signals alone reaches the
accuracy of the same estimator trained on the full real training set.
The result holds across three Pipelines, two downstream architectures,
and both sensors, and real windows can be exchanged for generated ones
up to complete replacement without measurable loss on the held-out real
trajectories. Downstream training is therefore no longer bound to the
quantity of recorded data, which is the constraint that sea trial cost
imposes on learned AUV navigation.

The rest of this paper is organized as follows.
Section~\ref{sec:background} presents the DVL measurement model.
Section~\ref{sec:approach} presents the diffusion framework, the
training procedure, the construction of the synthetic dataset, and the
evaluation protocol at both levels. Section~\ref{sec:results} reports
the signal-level and downstream results for the three Pipelines.
Section~\ref{sec:conclusion} concludes the paper.

\section{Background}
\label{sec:background}

\subsection{DVL Measurements}
\label{sec:measurements}

The DVL estimates velocity by transmitting acoustic pulses toward the seafloor and
measuring the Doppler shift of the returning echoes~\cite{brokloff1994}. Standard DVLs consist of four transducers. The most typical arrangement of these transducers
is the Janus configuration, in which four beams are spaced $90^{\circ}$ apart in azimuth
and tilted by a fixed angle $\alpha$ from the $z$-axis of the instrument, forming two
opposing pairs. The direction of the $i$-th beam is then~\cite{liu2018}: 
\begin{equation}
\begin{split}
    \mathbf{b}_i &=
    \begin{bmatrix}
        \cos \psi_i \sin \alpha & \sin \psi_i \sin \alpha & \cos \alpha
    \end{bmatrix}, \\
    \psi_i &= \frac{(i-1)\pi}{2} + \frac{\pi}{4},
     \quad i = 1, 2, 3, 4
\end{split}
\label{eq:beam_directions}
\end{equation}
Stacking the four directions into the transformation matrix
$\mathbf{H}$ relates the vehicle velocity to the beam velocity measurements
$\boldsymbol{\upsilon}_{\text{beams}}$~:
\begin{equation}
    \boldsymbol{\upsilon}_{\text{beams}} = \mathbf{H} \mathbf{v}^{d}, \quad
    \mathbf{H} =
    \begin{bmatrix}
        \mathbf{b}_1 \\
        \mathbf{b}_2 \\
        \mathbf{b}_3 \\
        \mathbf{b}_4
    \end{bmatrix}
    \in \mathbb{R}^{4 \times 3}
    \label{eq:beam_projection}
\end{equation}

where $\mathbf{v}^{d}$ is the velocity vector of the vehicle expressed in the DVL frame. Using a known transformation matrix from the DVL frame to the body frame, $\mathbf{T}^{b}_{d}$, the vehicle velocity vector expressed in the body frame is:

\begin{equation}
    \mathbf{v}^{b} = \mathbf{T}^{b}_{d} \mathbf{v}^{d}
    \label{eq:body_to_dvl}
\end{equation}

\section{Proposed Approach}
\label{sec:approach}

Deep learning approaches have been adopted across the DVL processing chain, offering data-driven alternatives to model-based velocity estimation. These networks require training data that covers the range of conditions they will encounter. For AUV navigation this data can consist of DVL and inertial recordings gathered during sea trials, which demand a dedicated platform, a reference sensor, and substantial ship time.

To avoid such a costly and time consuming procedure, we propose a conditional diffusion framework for generating synthetic DVL and inertial sensor signals. 

\subsection{Overview}
\label{sec:overview}
\begin{figure*}[t]
    \centering
    \includegraphics[width=\textwidth]{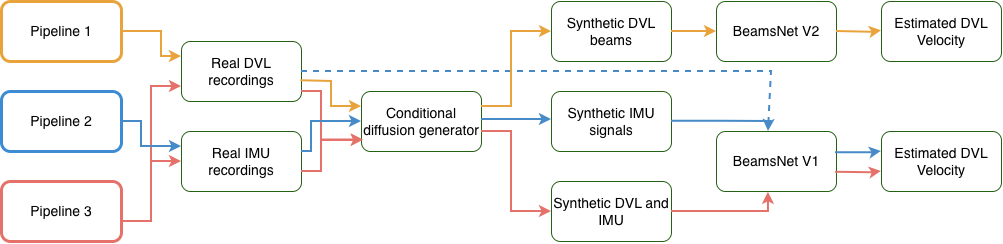}
    \caption{The three Pipeline configurations. Each differs in which sensor
signals are synthetic and which BeamsNet variant is trained downstream. Pipeline 1 generates the DVL beams alone and trains BeamsNetV2. Pipeline 2 generates the
inertial channels alone - the dashed line marks the real DVL beams, which are supplied to BeamsNetV1 alongside the generated inertial signals. Pipeline 3 generates the DVL beams and the inertial channels jointly from a single model and trains BeamsNetV1. A separate generator is trained for each pipeline.}
    \label{fig:pipelines_overview}
\end{figure*}

Our proposed approach employs Elucidated diffusion models (EDM), conditioned on descriptors of real signal recordings, to generate synthetic sensor signals that used to train an existing deep learning model. The EDM Pipeline follows Cohen and Klein~\cite{cohen2026diffusion}, who applied conditional diffusion only to synthetic inertial data for a downstream regression task, and we adopt their preconditioning, noise schedule and sampler. Our Pipeline offers a generative approach in which the signal characteristics of the inertial and DVL are learned from the recordings themselves. The generator is given recorded windows and descriptors computed from them, which serve as the conditions, and receives no DVL beam geometry, no mounting rotation, and no error parameters. As such, we create synthetic signals that can be used in place of real ones for training the downstream model (BeamsNet), reducing the dependence on extensive and costly sea trial data collection. 
We evaluate our proposed approach using three configurations, which differ in which sensor signals are synthetic and which BeamsNet version is used for evaluation, as illustrated in Figurre \ref{fig:pipelines_overview}. The two BeamsNet variants differ in their inputs. BeamsNetV1 estimates the DVL velocity from the beam measurements together with the inertial readings where BeamsNetV2 estimates it from the beam measurements alone.

The generated signals are assessed in two complementary ways. The downstream evaluation establishes whether the synthetic signals are good enough for training, but since its outcome depends on the downstream model as well as on the signals, the intrinsic metrics are needed alongside it. The intrinsic evaluation describes properties of the signals themselves but does not establish that they are good enough to support training a model. In the following, we describe our diffusion model Pipeline for generating the synthetic dataset.

\subsection{Diffusion Model}
\label{sec:diffusion_background}

AUV motion is bounded by the vehicle dynamics. The velocity, orientation, and turn rates are set by the trajectory of the mission, meaning the
DVL and the IMU produce only a narrow range of the signals they are
capable of measuring. A recorded trajectory is one realization of that
motion among many, since the currents, the exact path taken, and the
sensor noise all differ from one execution to the next. The recorded
trajectories are therefore a finite sample of a much larger set of
recordings the vehicle could potentially produce, and each recorded signal is treated as a sample from a distribution $p(y)$ over that set ~\cite{fefferman2016testing}. The objective of the generator is to draw a further sample from $p(y)$, one that was not among the recordings collected, and to draw it for a specified motion rather
than an arbitrary one. 

Supplying the motion descriptors, which are defined in Section~\ref{sec:preliminaries}, and collected into a
conditioning vector $c$, replaces that draw with one from the
conditional distribution $p(y \mid c)$, so that the generated signal
describes the motion $c$ specifies.

Two processes are involved in sampling from $p(y)$. In the forward
process, the recorded signals are corrupted by adding zero-mean Gaussian white noise, and a denoiser is trained to recover the clean signal
from the noisy one~\cite{sohl2015deep}. We follow an EDM
formulation~\cite{karras2022edm}, in which the corruption is
indexed by a continuous noise level $\sigma$, expressed in the signal
units, defined by:
\begin{equation}
  x = y + n , \qquad n \sim \mathcal{N}\!\left(0, \sigma^{2} I\right)
  \label{eq:corruption}
\end{equation}
where $y$ is a recorded signal, $n$ is zero-mean Gaussian white noise with standard
deviation $\sigma$, and $x$ is the noisy signal. The reverse process inverts this and draws the sample, that is, generates the signal. It starts at a noise level high enough that $n$ dominates $y$ in
\eqref{eq:corruption}, so that the corrupted signal is
indistinguishable from pure noise, and applies the denoiser over a
decreasing sequence of levels, down to one at which the remaining
noise is negligible relative to the signal. Since no recorded signal
lies behind the initial noise, the result is a signal that was never
recorded. At each level the denoiser produces an estimate of the clean signal, and that estimate sets the direction of a step to the next, lower level, where it is recomputed. Repeating this down the sequence turns the initial noise into a signal.

The conditioning vector shifts the generation toward the signals
consistent with a specified motion. Without it the model draws from
$p(y)$ alone. The generated signal is then whatever the initial noise
leads to, and the motion it describes cannot be chosen. In training,
the conditioning vector is computed from the clean recorded signal
before that signal is corrupted, so it always describes what the
denoiser is recovering. The denoiser receives the corrupted signal,
the noise level and the conditioning vector together, and each
recorded signal therefore contributes the triplet $(x, y, c)$, which
is what reaches the conditional distribution
$p(y \mid c)$~\cite{dhariwal2021}. At generation the
conditioning vector is set on its own, with no recorded signal behind
it, and since a mean, a STD and a peak map do not pin down a single
signal, the model is free to return one that was never recorded.

The reverse process applies a denoiser $D$ at each noise level to
produce an estimate of the clean signal. It is built around
$F_\theta$, a trainable network whose architecture is given below, and
whose input and output are scaled by $\sigma$. Following the EDM
preconditioning~\cite{karras2022edm}, a fixed set of
noise-level-dependent scalings applied around the network:
\begin{equation}
  D(x; \sigma, c) = c_{\text{skip}}(\sigma)\, x
  + c_{\text{out}}(\sigma)\,
    F_\theta\!\left(c_{\text{in}}(\sigma)\, x;\,
                    c_{\text{noise}}(\sigma),\, c\right)
  \label{eq:denoiser}
\end{equation}
where $x$ is the corrupted signal of \eqref{eq:corruption} and $c$ is
the conditioning vector. The input and output scalings are
\begin{equation}
  c_{\text{in}}(\sigma) =
    \frac{1}{\sqrt{\sigma^{2} + \sigma_{\text{data}}^{2}}} ,
  \qquad
  c_{\text{out}}(\sigma) =
    \frac{\sigma\, \sigma_{\text{data}}}
         {\sqrt{\sigma^{2} + \sigma_{\text{data}}^{2}}}
  \label{eq:precond_inout}
\end{equation}
and the skip and noise scalings are
\begin{equation}
  c_{\text{skip}}(\sigma) =
    \frac{\sigma_{\text{data}}^{2}}
         {\sigma^{2} + \sigma_{\text{data}}^{2}} ,
  \qquad
  c_{\text{noise}}(\sigma) = \frac{\ln \sigma}{4}
  \label{eq:precond_skipnoise}
\end{equation}
where $c_{\text{in}}$ normalizes the input amplitude so that the
network sees inputs of roughly unit variance across noise levels,
$c_{\text{out}}$ scales the network output, $c_{\text{skip}}$ sets how
much of the corrupted signal is passed directly to the output, and
$c_{\text{noise}}$ maps the noise level to the scalar fed to the
noise-level embedding. The constant $\sigma_{\text{data}}$ is the
reference scale of the EDM preconditioning, set per configuration, as described in Section~\ref{sec:syn_data_generation}.

\begin{figure*}[t]
    \centering
    \includegraphics[width=\linewidth]{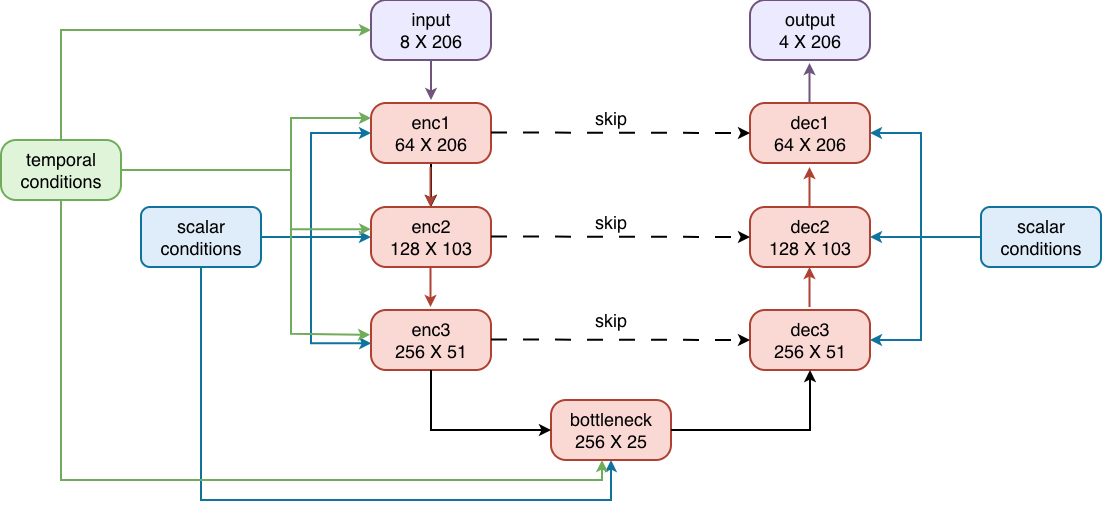}
    \caption{Structure of the conditional diffusion U-Net, shown for the beams-only
configuration (BeamsNet V2). The four beam channels and their four peak-map channels enter
together as eight input channels, and the network predicts the four beam channels
alone. The scalar conditions modulate
every residual block. The temporal condition is injected on the encoder side and
at the bottleneck, and reaches the decoder through the skip connections.}
    \label{fig:diffusion_overview}
\end{figure*}

The backbone $F_{\theta}$ is a one-dimensional U-Net
(Figure~\ref{fig:diffusion_overview}). It receives the $N_{\text{ch}}$
noisy signal channels together with their $N_{\text{ch}}$ peak-map
channels and outputs $N_{\text{ch}}$ channels, which the preconditioning
of $D_{\theta}$ maps to the denoised signal. $N_{\text{ch}}$ is
four, six, or ten for Pipelines~1, 2, and~3, respectively. The encoder
has three stages, each halving the signal length by average pooling,
followed by a bottleneck at the coarsest resolution. The decoder mirrors the encoder with nearest-neighbour
upsampling and restores the original length, and skip connections
concatenate each encoder stage with the decoder stage at the same
resolution. Each stage is a residual block of two convolutions with
group normalization and SiLU activations. The base width is 64
channels when the beams are generated alone and 96 channels when
inertial channels are present, with condition embeddings of 64 and 128
dimensions, respectively.

The conditioning is built from three per-channel descriptors of the
vehicle motion: the mean, the standard deviation (STD), and a peak map
that marks the position, sign, and strength of maneuver events in a
window. Their definitions and the reasoning behind the choice are given
in Section~\ref{sec:preliminaries}, and the equations are referred to
here as needed.
The conditioning splits into two: the scalar conditions, which are the
signal mean and STD, and the temporal condition, the peak map, which
is of the same length as the signal and cannot be represented as a
per-channel scalar. They enter the network by different mechanisms,
since they differ in form. The scalar conditions describe global
properties of the signal, and are each mapped to an embedding and
injected into every residual block, modulating the features after each
normalization layer through FiLM~\cite{perez2018film}:
\begin{equation}
  \mathrm{FiLM}(h, c) = \left(1 + \gamma(c)\right) h + \beta(c)
  \label{eq:film}
\end{equation}
where $h$ is a feature map and $\gamma(c)$, $\beta(c)$ are a
per-channel scale and shift predicted from the condition embedding.
The noise level is injected as an additive per-channel shift in the
same blocks. The peak map is concatenated with the generated
channels at the network input and added as a feature map at each
encoder stage and at the bottleneck, being linearly resampled to each
resolution beforehand. It is not injected on the decoder side, which
receives it through the skip connections.

In the joint configuration (Pipeline 3) the beam channels and the
inertial channels are given separate conditioning pathways. A shared
pathway would let a single projection layer predict the scale and
shift of a beam channel in part from the inertial conditions, and
those of an inertial channel in part from the beam conditions. During
training the two sets of conditions are computed from the same
recorded epoch and are therefore correlated, so a shared layer can
exploit that correlation to lower the training loss. At generation,
however, the conditions are supplied as separate design inputs and can
be combined in ways that never occurred in the recordings, and a
generator that relies on the correlation has no basis for such
combinations. Separate pathways ensure that the scale and shift
applied to a beam channel are computed from the beam conditions alone.
 
\subsection{Training Process}
\label{sec:training_process}
Training carries out the forward process of \eqref{eq:corruption} on
the recorded windows of the training trajectories. At each step a
window is corrupted at a noise level drawn at random, and the denoiser
is updated on how well it recovers the original. The clean sample $y$
of \eqref{eq:corruption} is a recorded window smoothed by a Gaussian
filter of width $s = 0.2$, so that the denoiser learns the vehicle
motion rather than the sensor noise carried in the recordings, at the
cost of some high-frequency content. Condition dropout \cite{ho2022cfg} is applied during training with the following rule: with probability
$0.2$ all conditions are dropped, the scalar embeddings being replaced
by a null embedding and the peak-map channels set to zero. The denoiser therefore also learns the unconditional task. Guidance is not used at generation, and the conditional prediction is taken directly. The training objective is the
noise-level-weighted reconstruction error:

\begin{equation}
\label{eq:error}
  \mathcal{L} = \mathbb{E}_{y, \sigma, n}
    \left[ \lambda(\sigma)
      \left\| D(x; \sigma, c) - y \right\|_{2}^{2} \right] 
\end{equation}

where $\mathcal{L}$ is the training loss, $D(x; \sigma, c)$ is the
denoiser output of \eqref{eq:denoiser}, $y$ is the clean signal it is
trained to recover, and $\lambda(\sigma)$ is the noise-level weight, defined by:

\begin{equation}
 \qquad
  \lambda(\sigma) =
    \frac{\sigma^{2} + \sigma_{\text{data}}^{2}}
         {\left(\sigma\, \sigma_{\text{data}}\right)^{2}}
  \label{eq:loss}
\end{equation}

The loss is averaged over the three quantities that vary between
training samples: the recorded signal $y$, the noise level $\sigma$,
and the noise draw $n$. Together these fix the corrupted signal $x$
through \eqref{eq:corruption}. The weight
$\lambda(\sigma)$ is intended to equalize the contribution of each
noise level to the loss. It grows without bound as $\sigma$ approaches
zero, and is capped at $10$ for numerical stability.

The loss of \eqref{eq:error} is minimized with Adam, at a learning rate
of $10^{-4}$ and weight decay $10^{-4}$, with gradient-norm clipping
at $1.0$. During training the noise level is drawn per sample as
$\ln \sigma \sim \mathcal{N}(P_{\text{mean}}, P_{\text{std}}^{2})$,
with $P_{\text{std}} = 1.2$ following~\cite{karras2022edm}. The beams-only
model (Pipeline 1) is trained for 7500 epochs at batch size 4 with a
fixed learning rate. The configurations with inertial channels
(Pipelines 2 and 3) are trained for 5000 epochs at batch size 32 with
cosine annealing from $10^{-4}$ to $10^{-6}$. The epoch counts are not
comparable across Pipelines, because the training sets differ in size:
Pipeline 1 draws from 100 windows and Pipelines 2 and 3 from 4000.
The training process is summarized in Algorithm~\ref{alg:training}.

\begin{algorithm}[t]
\caption{Denoiser Training}
\label{alg:training}
\begin{algorithmic}[1]
\Statex \textbf{Input:} Recorded windows of the training trajectories:
network $F_{\theta}$, number of epochs $E$
\Repeat
  \State Draw a recorded window and smooth it with a Gaussian filter of
         width $s = 0.2$ to obtain the clean signal $y$
  \State Compute the mean, STD and peak map of $y$ by
         (\ref{eq:cond_mean})--(\ref{eq:peak_map}) and collect them into
         the conditioning vector $c$
  \State With probability $0.2$, drop all conditions: replace the scalar
         embeddings by a null embedding and set the peak-map channels to
         zero
  \State Draw a noise level $\sigma$ from the log-normal distribution
         $\ln \sigma \sim \mathcal{N}(P_{\text{mean}},
         P_{\text{std}}^2)$
  \State Corrupt the clean signal at that level to obtain the noisy
         signal $x$ of (\ref{eq:corruption})
  \State Evaluate the denoiser $D(x; \sigma, c)$ of (\ref{eq:denoiser})
  \State Compute the loss of (\ref{eq:error}) with the weight
         $\lambda(\sigma)$ of (\ref{eq:loss})
  \State Update $\theta$ with Adam
\Until{$E$ epochs are completed}
\State \textbf{return} Trained denoiser $D$
\end{algorithmic}
\end{algorithm}

\subsection{Synthetic Data Generation}
\label{sec:syn_data_generation}

The synthetic data is built from the A-KIT recordings~\cite{akit}, which provide the reference velocity, the DVL measurements, and the IMU measurements for thirteen trajectories, each of duration 400 seconds, of the AUV. For each trajectory, the reference velocity
is mapped to the DVL frame by \eqref{eq:body_to_dvl}, and the result is
projected onto the four beam directions through the geometry matrix
$\mathbf{H}$ of \eqref{eq:beam_projection} to obtain the four beam velocities.
The IMU measures specific force and angular velocity, vectors expressed in the body frame, in three accelerometer channels and three gyroscope channels, and is rotated to the DVL frame by the same mounting rotation. The rotation is required because the beam geometry matrix is defined in the DVL frame, and the inertial channels are rotated to the same frame so that all generated channels share a common basis.
 
Pipelines 2 and 3 use inertial channels, and the two sensors are sampled at different rates. The DVL samples at 1 Hz and the IMU at 100 Hz, so each DVL epoch is paired with the 100 inertial samples.
 
Window length follows the sensor type. The beam channels of Pipeline 1 are sliced into windows of 206 samples with a stride of 20 samples,
yielding ten windows per trajectory. The inertial windows of
Pipelines 2 and 3 are the 100-sample blocks, one per DVL epoch.
 
Channel normalization depends on whether the configuration mixes
sensors. When the beams are generated alone (Pipeline 1), they are
used in raw physical units. Their standard deviations lie between 0.11
and 0.14 across the four channels, so the channels are already on a
common scale and no normalization is applied. When inertial channels
are generated, alone (Pipeline 2) or jointly with the beams (Pipeline
3), each channel is standardized to zero mean and unit variance using
a fixed mean and standard deviation computed once over the training
set, and generated signals are returned to physical units by the
inverse transform. Standardization is required in these configurations
because the channels differ in physical units and in scale by roughly
an order of magnitude relative to one another. Without it, the smaller
channels are corrupted more heavily than the larger ones at every
noise level.
 
The two preconditioning constants of Section~\ref{sec:diffusion_background} follow from this choice. The constant $\sigma_{\text{data}}$ is set to
0.5 for the un-normalized beams, following the EDM reference
value~\cite{karras2022edm}, and to 1.0 for the standardized channels,
matching their unit variance. The preconditioning of
\eqref{eq:precond_inout} and \eqref{eq:precond_skipnoise} depends on
$\sigma$ only through the ratio $\sigma / \sigma_{\text{data}}$, and
$P_{\text{mean}}$ is the mean of $\ln \sigma$, so $P_{\text{mean}} =
-1.2$ is used for the beams and $-1.2 + \ln 2$ for the standardized
channels, which leaves the sampled range of that ratio the same in
both.

The three motion descriptors of Section ~\ref{sec:preliminaries} are computed per channel from the same windows. Two peak-detection parameters scale with the window length. For the 206-sample beam windows the trend window is 50 samples and the local half-width is 20 samples. For the 100-sample inertial windows both are halved, since a trend window covering half the signal would suppress the information the descriptor is meant to locate. The variability threshold is 0.4 in all configurations.

After the channels are prepared and the descriptors are computed, the synthetic dataset can be generated. The previous stage produced one trained denoiser per Pipeline. Synthetic windows are produced by running the reverse process as described in Section~\ref{sec:diffusion_background}. A sample is drawn from
$\mathcal{N}(0, \sigma_{\max}^{2} I)$ with $\sigma_{\max} = 80$, and
the reverse process is integrated over a log-spaced schedule of 200
levels down to $\sigma_{\min} = 0.002$, following the EDM
values and using the deterministic second-order Heun
sampler of~\cite{karras2022edm}. The conditioning vector is supplied
at every level. The sampler is deterministic, so the variation between
draws from one conditioning vector comes from the initial noise alone.
Each pass produces one
window, and a synthetic dataset is built by repeating the process with
a different conditioning vector each time.

The design challenge at this stage is which conditioning vectors to
supply. A window generated from the descriptors of a real recording
may produce a regime already present in the data and adds little. The
conditioning vectors are therefore constructed to describe motion that
the AUV could have performed but did not. Such motion belongs to the
set of recordings the vehicle could produce, since the vehicle is
capable of it, but was not sampled by the recordings. The construction
must also keep the descriptors within the range the denoiser was
trained on. Each generated window is conditioned either on a pair of
windows or on a single real window.
 
Pipeline 1 uses the paired method. It generates the beams alone and
carries its own label, so nothing in the output has to correspond to a
particular recorded epoch and the conditioning vector is free to
describe one that was never recorded. A second window is drawn from
the same trajectory, within five window positions of the first, and
the two signals are blended:
\begin{equation}
  y_{ab} = \kappa\, y_a + (1 - \kappa)\, y_b , \qquad
  \kappa \sim \mathrm{Beta}(0.5, 0.5)
  \label{eq:blend}
\end{equation}
where $y_a$ and $y_b$ are the two real windows, $\kappa$ is the
blending weight, and $y_{ab}$ is the blended window. 

Interpolating between two conditions by a combination has been used in conditional generation to reach intermediate regimes~\cite{alcaraz2023diffusion}. Here, the descriptors are computed from the blended signal
rather than by blending the descriptors themselves. The blend
describes a motion between two windows, which the AUV could have
performed but did not. The two windows are drawn close together in the
same trajectory so that the blended descriptors stay within the region
the denoiser was trained on. Pairing across trajectories or at
arbitrary separation results in unrealistic trajectories. The STD and
the peak map are not linear, and computing them from the blended
signal ensures the three descriptors describe a signal that exists, as
they do for a real window.
 
Pipeline 2 uses the single scheme. It generates only the inertial
signals and uses the real beams and the real reference velocity of the
source epoch, so the conditioning vector must describe that specific
epoch and the descriptors of one real window cannot be changed.
 
Pipeline 3 uses both schemes. It generates all ten channels, so a
blended draw needs no real counterpart and the same pair and weight
can be applied across every channel. One draw per epoch is taken on
that epoch's own descriptors and five further draws on blended pairs.

The downstream model described in Section~\ref{sec:downstram_eval}
requires labels, therefore a label is produced alongside each synthetic
window. Each label traces back to the real recordings. It is the
reference velocity of the source velocity, or, for a blended pair, the
same blend of two reference velocities. No label is derived from a generated signal at any point, which would otherwise make the downstream training set self-referential. The generation process is summarized in Algorithm~\ref{alg:generation}. The generated signals are evaluated at two levels. The signal-level metrics describe properties of the generated windows themselves. The two levels look at the generated signals from different angles. A set of signals can satisfy every signal-level metric and still fail to train a model. Similarly, a model can train successfully on signals whose properties have not been characterized. The two levels are therefore complementary, and together
they give the full picture.

\begin{algorithm}[t]
\caption{Synthetic Data Generation}
\label{alg:generation}
\begin{algorithmic}[1]
\Statex \textbf{Input:} Trained denoiser $D$, real training windows,
number of synthetic windows $N$
\Statex \textbf{Sampler:} Heun, 200 log-spaced levels from
$\sigma_{\max} = 80$ to $\sigma_{\min} = 0.002$
\For{$n = 1$ to $N$}
  \State Draw a real window $y_a$ from a training trajectory
  \If{the draw is blended}
    \State Draw a second window $y_b$ 
    \State Blend $y_a$ and $y_b$ into $y_{ab}$ by (\ref{eq:blend}) and
           set $y \gets y_{ab}$
  \Else
    \State Set $y \gets y_a$
  \EndIf
  \State Compute the mean, STD and peak map of $y$ by
         (\ref{eq:cond_mean})--(\ref{eq:peak_map}) and collect them into
         the conditioning vector $c$
  \State Draw the initial noise $x_{\sigma_{\max}} \sim
         \mathcal{N}(0, \sigma_{\max}^2 I)$
  \State Integrate the reverse process from $\sigma_{\max}$ to
         $\sigma_{\min}$ with the Heun sampler and $D(\cdot; \sigma, c)$
         of (\ref{eq:denoiser}) to obtain the generated window
  \If{the channels were standardized}
    \State Return the generated window to physical units by the inverse
           transform
  \EndIf
  \If{the draw is blended}
    \State Assign the label as the same blend of the two reference
           velocities
  \Else
    \State Assign the label as the reference velocity of the source
           epoch
  \EndIf
\EndFor
\State \textbf{return} $N$ synthetic windows with their labels
\end{algorithmic}
\end{algorithm}

\subsection{Signal-Level Metrics}
\label{sec:signal_level_metrics}

We consider four signal-level metrics:
\begin{enumerate}
    
    \item \textbf{Condition adherence}: The objective is to establish that
    the conditioning input reaches the output, since a generator can learn
    the data distribution and ignore the condition. The per-channel mean of the conditioning vector is used as the probe, since it is the one
    descriptor that can be recomputed from a generated window and compared
    directly against the value that was requested. A window is generated
    from a condition whose mean component for channel $j$ is
    $\mu^{\text{target}}_{j}$, and the mean of the generated window
    $\hat{\mu}_{j}$ is computed by (\ref{eq:cond_mean}). The Pearson
    correlation between the two is reported per channel over a set of
    windows. A generator that ignores the conditioning produces a mean
    unrelated to the target and a correlation near zero.
    \item \textbf{Held-out conditioning transfer}: The objective is to establish that the conditioning mechanism operates on trajectories outside the set the generator was trained on. A generator that had fitted only the conditions present in its training set would return, for any other condition, a window drawn from the
    trajectories it was trained on. Such a window remains an acceptable sensor signal at approximately the requested level and amplitude, but it does not follow the signal the condition was computed from. The conditions are computed from windows of the test trajectories, which are withheld from generator training. A window is generated from each such condition and compared against the real window the condition was computed from, channel by channel. It is important
    to note that the test measures transfer of the conditions and not novelty - this is addressed separately by the fourth test. A generated signal that satisfies the requested descriptors and still departs from the real one indicates a limit of the conditioning.
    \item \textbf{Draw-to-draw diversity}: The objective is to establish that the conditioning leaves the generator
    enough freedom to produce different signals from the same condition. Diversity is treated as a dimension of generative model quality independent of
    fidelity \cite{alaa2022faithful}. The measurement here is conditional, taken
    across draws from a single condition, rather than across a full sample set. The
    descriptors of Section~\ref{sec:preliminaries} were chosen to fix the
    dynamic regime of a window while leaving its fine structure to the model. A condition that determined the
    window would leave no freedom, and repeated draws would return the same signal. The test establishes that the draws differ from one another. For this test, six windows are generated
    from each condition, each starting from an independent noise draw. The spread across
    draws is reported relative to the local standard deviation of the real
    window the condition was computed from. A generator whose output was fixed by the condition would give zero spread.
    \item \textbf{Nearest-neighbor replication}: This test follows the
    procedure of \cite{alaa2022faithful}. The objective is to establish that no generated
    window reproduces a window from the training set. This property is required because the value of a synthetic window rests on it being a new sample and
    not a copy. Generalization in this sense is independent of fidelity and diversity, since a model can score well on both by resampling its training data. For each synthetic window the $L_2$ distance to the closest
    window in the real training set is computed per channel. The resulting distribution is read against a leave-one-out reference, in which a each real training window is measured against all other training windows excluding
    itself. A generated window that reproduced a training window would lie at or near zero distance from it, since the reproduced window is then its own nearest neighbor. 
\end{enumerate}

\subsection{Downstream Evaluation}
\label{sec:downstram_eval}
The downstream model is BeamsNet \cite{beamsnet}. It is an end-to-end
deep learning framework that regresses the DVL velocity vector from the beam
measurements, proposed to replace the model-based least squares estimator, which is the classical solution for recovering the DVL velocity vector from the four beam velocity measurements. BeamsNet leverages deep learning capabilities in order to capture nonlinear behaviour and other uncertainty properties in the data. Two versions are proposed, differing in their input to the network.
\begin{enumerate}
    \item \textbf{BeamsNetV1}: This model employs the DVL beam measurements of a single epoch together with
the inertial sensor readings to regress the DVL velocity vector of that epoch
using a 1DCNN. To cope with the different input sizes, its architecture contains three heads. The accelerometer and gyroscope measurements pass through convolutional layers that extract features from the data, and the extracted features are flattened and passed through a sequence of fully connected layers. The last head takes the DVL beam measurements, which are combined before the last fully connected layer that produces the $3 \times 1$ estimated DVL velocity vector. 
    \item  \textbf{BeamsNetV2}: This model uses the beam measurements of $n$ past epochs in addition to those of the estimated epoch, and requires no IMU. The network is a two-headed 1DCNN, and the beam measurement of the estimated epoch is again combined before the last fully connected layer.

\end{enumerate}

The architecture of both versions is that of \cite{beamsnet} and is
unchanged. The training optimization uses Adam with a learning rate of $10^{-3}$. BeamsNetV2 is trained at batch size 4 and BeamsNetV1 at
batch size 32. 
The velocity that BeamsNet estimates is defined by the beam error model. Two
DVL units would be required to obtain an independent reference, one serving as
the unit under test and the other as ground truth (GT), and only one is available.
Following the procedure in \cite{beamsnet}, the velocity reported by the DVL is therefore
treated as GT, and the beams supplied to the network are produced from it. It is mapped to the DVL frame by (\ref{eq:body_to_dvl}) and projected onto the four beam directions through
$\mathbf{H}$ of (\ref{eq:beam_projection}). Each beam velocity is then corrupted by a bias,
a scale factor, and zero-mean white Gaussian noise, giving the beam error
model \cite{tal2017}:
\begin{equation}
\tilde{\boldsymbol{\upsilon}}_{\mathrm{beams}}
  = \bigl[\mathbf{H}\mathbf{v}^d(1+\mathbf{s})\bigr] + \mathbf{b} + \boldsymbol{\eta}
\label{eq:errmodel}
\end{equation}

where $\mathbf{s}$ is the DVL scale factor, $\mathbf{b}$ is the
DVL bias, and $\boldsymbol{\eta}$ is the zero-mean white Gaussian noise. The bias, the scale factor and the noise
standard deviation are taken from \cite{beamsnet}. The network therefore
receives corrupted beam velocities and is trained to recover the velocity they were
projected from. 
The thirteen A-KIT trajectories are split into ten for training and three for
testing. The three test trajectories are held out from the diffusion model
and from BeamsNet, so no generated window used for downstream training
derives from a test trajectory, and the conditioning statistics are taken from trajectories the generator never
saw. The same split is used in all three Pipelines and in every configuration
reported, so all results are read against the same test set. The test
trajectories were selected to span the range of motion present in the
recordings.
Accuracy is reported as the velocity RMSE on the held-out real trajectories:
\begin{equation}
\mathrm{RMSE} = \sqrt{\frac{1}{N}\sum_{k=1}^{N}
                \bigl\|\mathbf{v}^d_k - \hat{\mathbf{v}}^d_k\bigr\|_2^2}
\label{eq:rmse}
\end{equation}
where $\mathbf{v}^d_k$ is the GT velocity vector of epoch $k$, $\hat{\mathbf{v}}^d_k$ is the
network estimated velocity vector, and $N$ is the number of test epochs.
In the original BeamsNet paper, the RMSE is used to compare estimators on a fixed dataset, and it
establishes the improvement of BeamsNet over the least squares solution.
In this paper, the estimator is held fixed instead and the training data is varied, so
the same metric compares datasets rather than models. The relative improvement is reported as:
\begin{equation}
Improvement = 100 \cdot \frac{\mathrm{RMSE}_{\mathrm{real}} - \mathrm{RMSE}}
{\mathrm{RMSE}_{\mathrm{real}}}
\label{eq:improvement}
\end{equation}

Four experiments, as described below, are run on this basis. Each varies the composition of the
BeamsNet training set. The network architecture, the training settings and the
test set are held fixed throughout, so any change in RMSE is attributable to
the training data alone.

The evaluations are: 
\begin{enumerate}
    \item \textbf{Fully synthetic sweep}: The training set consists of
generated windows only, and no real window is used. The real training pool is still the reference against which the amount of synthetic data is measured. A ratio of 1:1 means the network is trained on as many synthetic windows as there are real windows available for training. A ratio of 5:1 means five times that number. The ratio is swept from 1:1 to 5:1. Accuracy is read against the real-only baseline, which is the same network trained on the real training pool alone. The objective is to establish whether generated recordings can stand in for real ones on their own, and whether enlarging the synthetic pool recovers the accuracy obtained from real recordings.
    \item \textbf{Combined synthetic and real sweep}: The training set
consists of the full real training pool, to which synthetic windows are added. The amount added is swept over the same range as in the first experiment, from 1:1 to 5:1 relative to the real training pool. Accuracy is read against the real-only baseline. The objective is to determine whether generated windows supplement real ones, which is the
setting in which a generator is used when recordings are available but limited in quantity.
    \item \textbf{Real-to-synthetic substitution sweep}: The total number of training windows is held constant. A fraction of the real training set is removed and replaced by an equal number of synthetic windows, and that fraction is swept from $0\%$ to  $100\%$. The objective is to isolate the effect of replacing real data from the effect of enlarging the training set. The sweep of the second experiment changes the pool size and its composition at the same time. Here the size is fixed, so the reported change measures how much accuracy is lost when a real window is exchanged for a generated one. The endpoint at $0\%$ is the real-only baseline. The endpoint at $100\%$ coincides with the smallest point of the fully synthetic sweep.
    \item \textbf{Conditioning ablation}:  The conditioning supplied to the
generator is reduced, and the resulting synthetic windows are used to train
BeamsNet at a 1:1 ratio to the real training pool. Three reduced
variants are evaluated against the full conditioning: the peak map is removed, the per-channel mean and STD are removed, and all conditioning
is removed. The downstream training protocol is identical in every variant, so the reported RMSE differences reflect the generator input alone. The objective is to measure the contribution of each conditioning component to the usefulness of the generated data. Reported change is relative to the full conditioning.
\end{enumerate}

Table \ref{tab:experiment_coverage} summarizes the four downstream experiments across the three
pipelines. We note that Pipeline 2 does not admit the full set of experiments. In this Pipeline only the inertial channels are generated (as can be seen in Figure \ref{fig:pipelines_overview}), and each synthetic IMU window is paired with a real DVL window. Every training window therefore contains real DVL beams by
construction, so a fully synthetic training set cannot be formed. The pairing is one to one, so the synthetic pool is bounded by the number of real DVL windows, and the second experiment covers 0:1 to 1:1 rather than 0:1 to 5:1.

\begin{table} 
\small
\caption{Coverage of the four downstream experiments across the three Pipelines (P1/P2/P3). The first two experiments and the ablation vary the size of the synthetic set against the real training set and are given as ratios. The substitution experiment holds the total number of training windows fixed and is given as the fraction of real windows replaced.}
\label{tab:experiment_coverage}
\begin{tabular}{|l|c|c|c|}
\hline
\textbf{Experiment} & \textbf{P1} & \textbf{P2} & \textbf{P3} \\
\hline
\makecell[l]{Fully synthetic\\sweep} & 1:1--5:1 & Not defined & 1:1--5:1 \\
\hline
\makecell[l]{Combined synthetic\\and real sweep} & 0:1--5:1 & 0:1--1:1 & 0:1--5:1 \\
\hline
\makecell[l]{Real-to-synthetic\\substitution} & 0--100\,\% & 0--100\,\% & 0--100\,\% \\
\hline
\makecell[l]{Conditioning\\ablation} & 1:1 & 1:1 & 1:1 \\
\hline
\end{tabular}
\end{table}

\section{Results}
\label{sec:results}
\subsection{Preliminaries}
\label{sec:preliminaries}
A conditional diffusion model generates a window from a description supplied at generation time. In order to conditionally generate a dataset, the dataset first has to be characterized by a dynamic specification that appears across trajectories and describes each one well. An effective condition is a low-dimensional summary that fixes the essential character of a window while leaving the fine detail to the model. A description that is too detailed constrains the signal almost completely, which collapses generation toward reproducing the conditioning window rather than sampling a regime. Low-order statistics avoid this: they capture the dominant structure of the motion in a few numbers and leave the model free to generate diverse signals consistent with them \cite{choi2021ilvr}. The mean and standard deviation are the two lowest-order such statistics, describing the amplitude of the motion, and they form the basis of the description. The same construction is applied to every channel, whether it carries a beam velocity, a specific force, or an angular rate, so that a single conditioning scheme covers both sensors.

The per-channel mean is:
\begin{equation}
\mu_j = \frac{1}{W}\sum_{k=1}^{W} s_j[k]
\label{eq:cond_mean}
\end{equation}

where $\mu_j$ is the mean of channel $j$, $W$ is the window length, and $s_j[k]$ is the value of the $j$-th generated channel at sample $k$. The generated channels are one of the four beam velocities, the three accelerometer axes, or the three gyroscope axes, depending on the configuration.

The per-channel STD is:
\begin{equation}
\sigma_j = \sqrt{\frac{1}{W}\sum_{k=1}^{W}\bigl(s_j[k]-\mu_j\bigr)^2}
\label{eq:cond_std}
\end{equation}
where $\sigma_j$ is the standard deviation of channel $j$.

The mean and standard deviation summarize the amplitude of the motion but do not encode when events occur within the window. Thus a window with a single abrupt maneuver and a window in which the same variation is spread uniformly might share the same $\mu_j$ and $\sigma_j$. For this reason, a third descriptor is needed to locate events in time. In this paper, a peak map is used. It marks the temporal position, sign, and strength of the maneuver events in a window. Events are detected on the detrended channel $\tilde{s}_j$, obtained by subtracting a moving average so that sharp excursions are exposed independently of slow baseline drift. Each detected event $q$ at sample $l_q$ contributes a signed Gaussian bump, and the peak map is their sum:
\begin{equation}
    m_j[k] = \sum_{q} \operatorname{sign}\bigl(\tilde{s}_j[l_q]\bigr)\,\rho_q
         \exp\!\left(-\frac{(k-l_q)^2}{2w_q^{2}}\right)
    \label{eq:peak_map}
\end{equation}

where $m_j[k]$ is the peak map of channel $j$ at sample $k$, $\tilde{s}_j$ is the detrended channel, $l_q$ is the sample at which event $q$ was detected, $\rho_q$ is the prominence of event $q$, the sign distinguishes upward from downward excursions,
and $w_q$ is a width that scales with the event's duration, small for sharp events and larger for broad ones. 

\begin{figure} 
    \centering
    \includegraphics[width=1\linewidth]{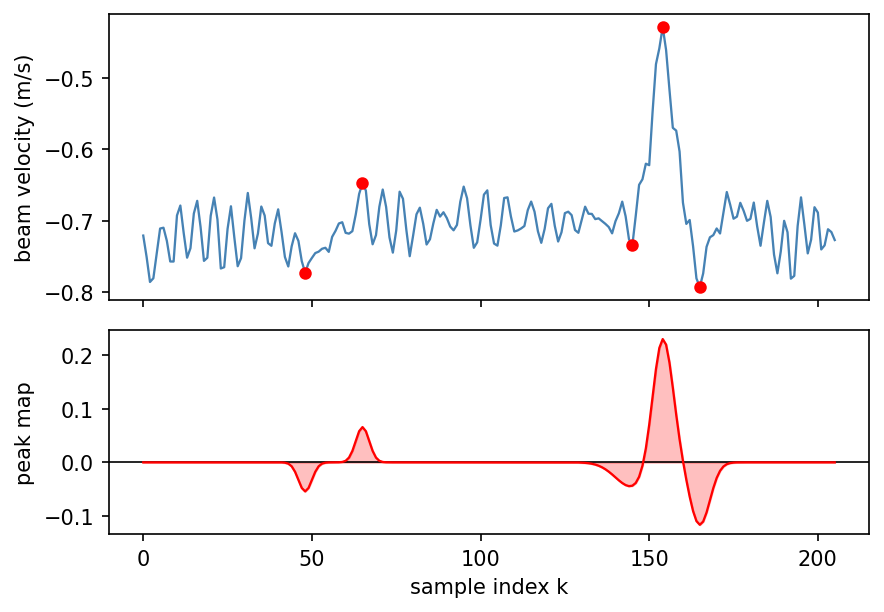}
    \caption{Example peak map for one DVL beam window. Top: the beam velocity
    signal with detected maneuver events marked. Bottom: the resulting peak map,
    a sum of signed Gaussian bumps placed at the detected events.}
    \label{fig:peak_example}
\end{figure}

The peak map therefore encodes where events occur, their direction, and their strength, while remaining sparse between events. Figure \ref{fig:peak_example} shows an example window, with the detected events on the beam signal and the resulting peak map. Together these three descriptors characterize the dynamic regime of a window, while still leaving the diffusion model enough freedom to generate diverse signals. 

\subsection{Signal-Level}
The four metrics of Section~\ref{sec:signal_level_metrics} are reported below. Each metric isolates one property of the generated signals, so a result on any one
of them points to a specific part of the Pipeline.
The conditioning design and the generation parameters were
developed on that basis. 

The generated channels are the four DVL beam velocities, the three
accelerometer specific-force components, and the three gyroscope angular
rates. In the figures, \text{beam1}--\text{beam4} denote the beam
velocities, \text{ax} the $x$ accelerometer component, and
\text{gx} and \text{gz} the $x$ and $z$ gyroscope components. For
clarity, a representative subset of channels is plotted in each figure but
all reported metrics are computed over the full set of channels.

\subsubsection{Condition Adherence}
\label{sec:adherence}
\begin{figure} 
    \centering
    \includegraphics[width=1\linewidth]{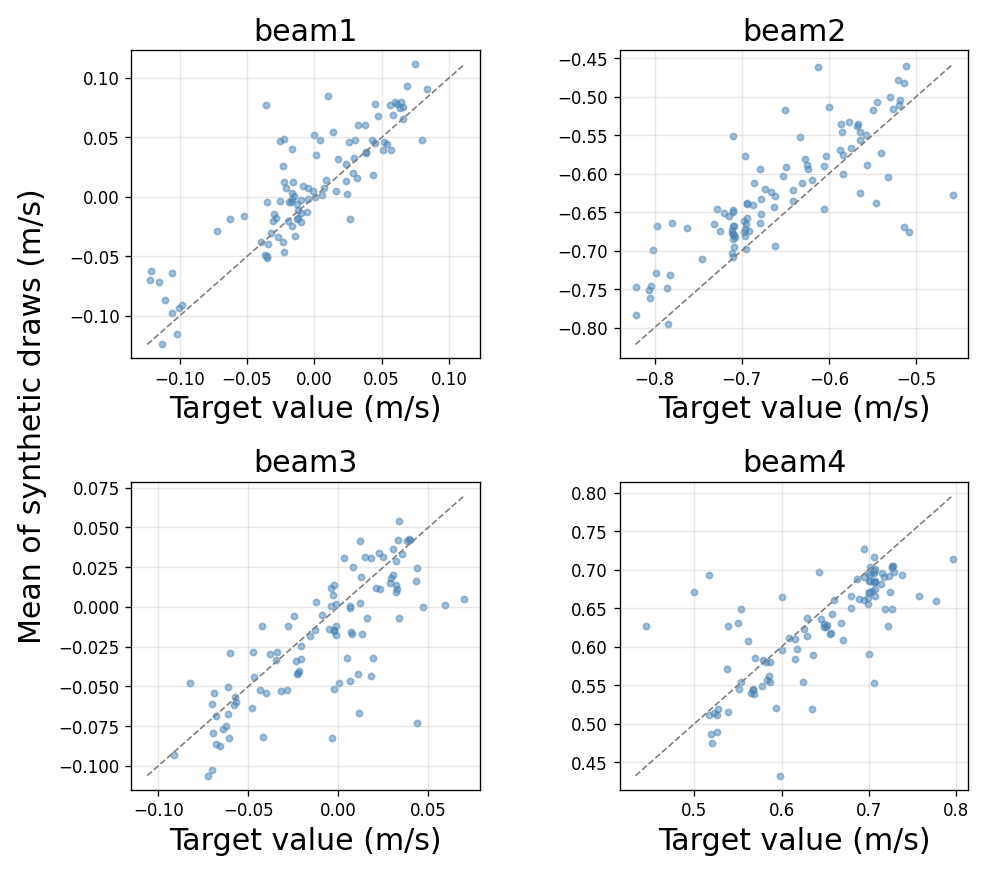}
    \caption{Pipeline 1, DVL signals, condition adherence test. All four DVL beam velocities are shown. The requested
    per-channel average is plotted against the average of the generated
    window, over 100 windows. The dashed line is the identity. The Pearson
    correlation between the requested and generated averages is 0.869 on
    \text{beam1}, 0.791 on \text{beam2}, 0.774 on \text{beam3}, and
    0.718 on \text{beam4}.}
    \label{fig:p1_adherence}
\end{figure}

\begin{figure} 
    \centering
    \includegraphics[width=1\linewidth]{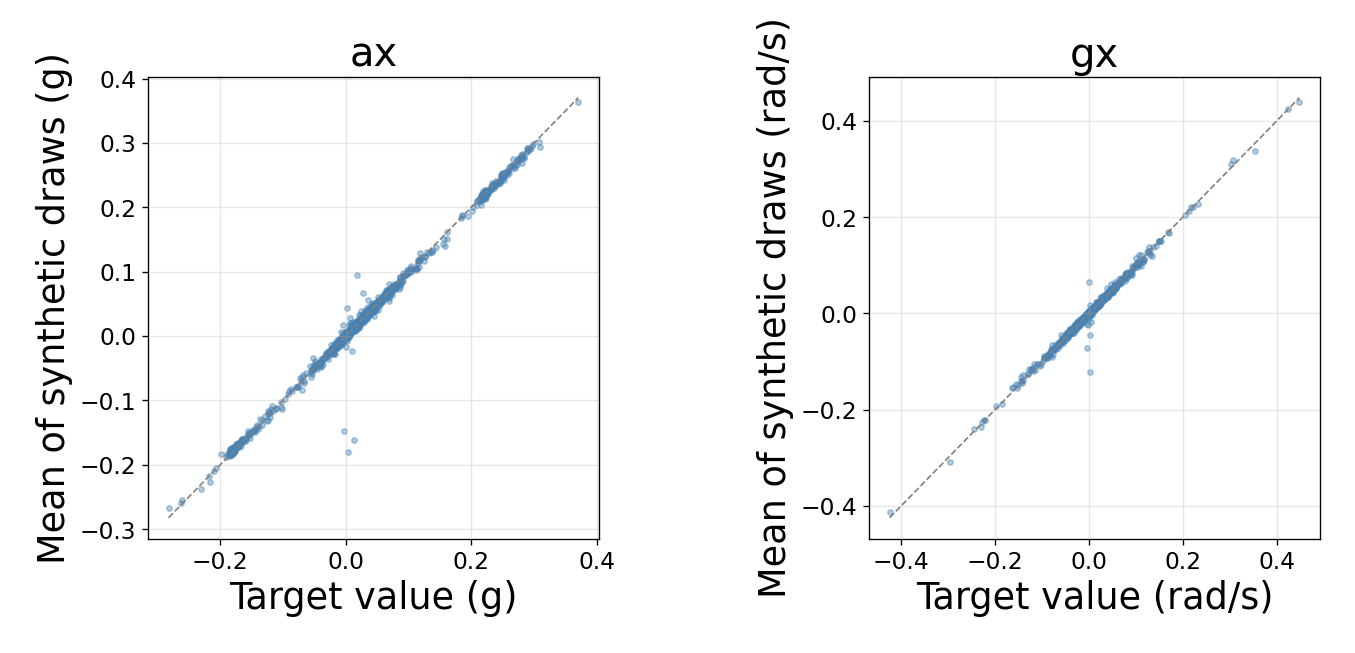}
    \caption{Pipeline 2, IMU signals, condition adherence test. Two
representative channels of the six are shown: the $x$ accelerometer
component (\text{ax}) and the $x$ gyroscope component (\text{gx}).
The requested per-channel average is plotted against the average of the
generated window. The dashed line is the identity. The Pearson
correlation between the requested and generated averages is 0.997 on
\text{ax} and 0.994 on \text{gx}.}
    \label{fig:p2_adherence}
\end{figure}

\begin{figure} 
    \centering
    \includegraphics[width=1\linewidth]{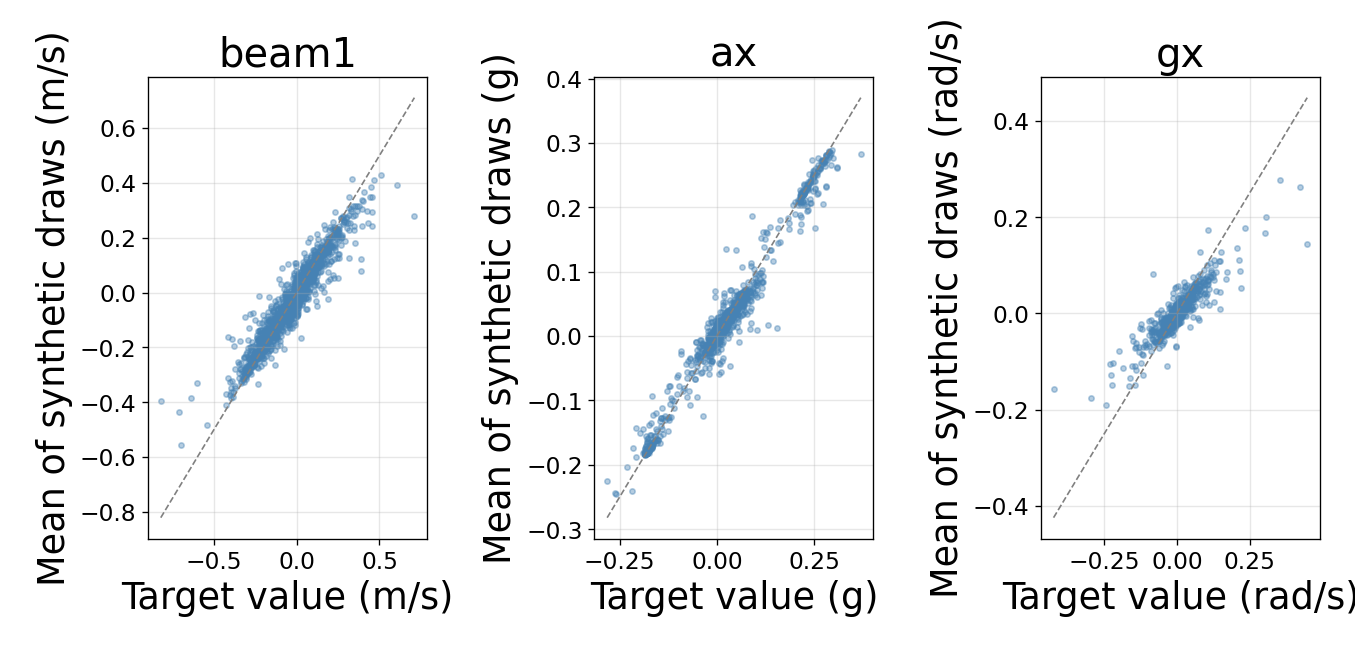}
    \caption{Pipeline 3, IMU and DVL signals, condition adherence test.
Three representative channels of the ten are shown, one per sensor type:
the first DVL beam velocity (\text{beam1}), the $x$ accelerometer
component (\text{ax}), and the $x$ gyroscope component (\text{gx}).
The requested per-channel average is plotted against the average of the
generated window. The dashed line is the identity. The Pearson
correlation between the requested and generated averages is 0.960 on
\text{beam1}, 0.988 on \text{ax}, and 0.895 on \text{gx}.}
    \label{fig:p3_adherence}
\end{figure}

The conditioning input reaches the output in all three Pipelines
(Figs.~\ref{fig:p1_adherence}, \ref{fig:p2_adherence} and
\ref{fig:p3_adherence}). No channel in any Pipeline shows a correlation near
zero, meaning the requested values are not ignored.

Agreement is highest on the accelerometer channels, followed by the beam
channels, and is lowest on the gyroscope channels. The ordering follows the
distribution of the requested values. Accelerometer averages are set primarily
by the projection of gravity onto the body axes, which varies with vehicle
orientation and is sustained over many windows. The requested values are therefore
distributed across the full range and the corresponding points spread along the
identity line. Beam averages are set by the vehicle velocity along the beam
directions, which is also sustained over extended intervals, so the beam
distribution is likewise broad. Gyroscope averages measure the angular rate,
which is near zero except during maneuvers, so the requested values are
concentrated near zero and the tails of the range are sparsely represented in
the training set.

The consequence is visible at the edges of the requested range, where the
points fall inside the identity line. The effect appears at the low end on all
four beam channels of Pipeline 1, and at both ends on beam3 and beam4.
Adherence is accurate in the populated region of the condition distribution and
regresses toward the marginal mean in the loosely sampled tails, while the
correlation remains high on every channel.

Pipeline 2 is the exception. Its gyroscope target distribution is concentrated
near zero as it is in Pipeline 3, but the tails are followed closely rather
than falling inside the identity line, and its correlations exceed the
corresponding Pipeline 3 values on both channels shown. A small number of
windows with targets near zero produce averages displaced from the target,
visible as isolated points off the identity line.

In Pipeline 3 the concentration in the center of the range is present on the gyroscope channels, which carry the lowest correlations of the sensors, and is also
visible on beam1 at both ends. Pipeline 3 takes most of its draws on blended conditions. The window mean is
linear in the signal, so the mean of a blended window is a combination of the two source means and cannot be more extreme than either. The blended conditions are therefore concentrated more centrally than the real ones.
The conditioning mechanism is therefore functional in all three Pipelines. Its accuracy varies with how densely the
requested region is represented in the training data, both across the range
within a Pipeline and across Pipelines, which differ in the number of training
windows available to the generator.

\subsubsection{Held-Out Conditioning Transfer}
\label{sec:results-transfer}

\begin{figure} 
    \centering
    \includegraphics[width=\linewidth]{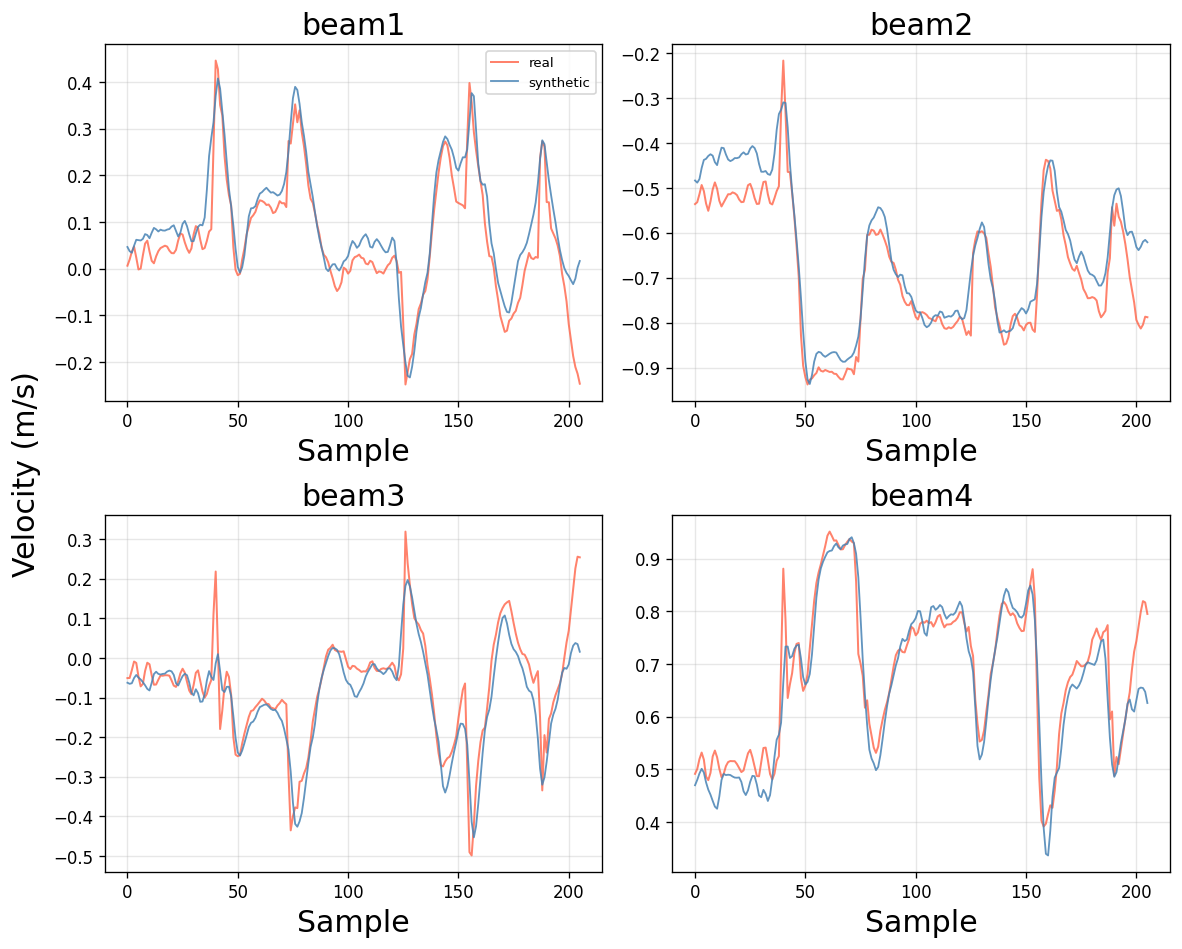}
    \caption{Pipeline 1, held-out conditioning transfer. All four beam
    channels are shown over a single 206-sample window. The conditioning
    statistics were taken from trajectory T8, withheld from training.}
    \label{fig:p1_transfer}
\end{figure}

\begin{figure} 
    \centering
    \includegraphics[width=\linewidth]{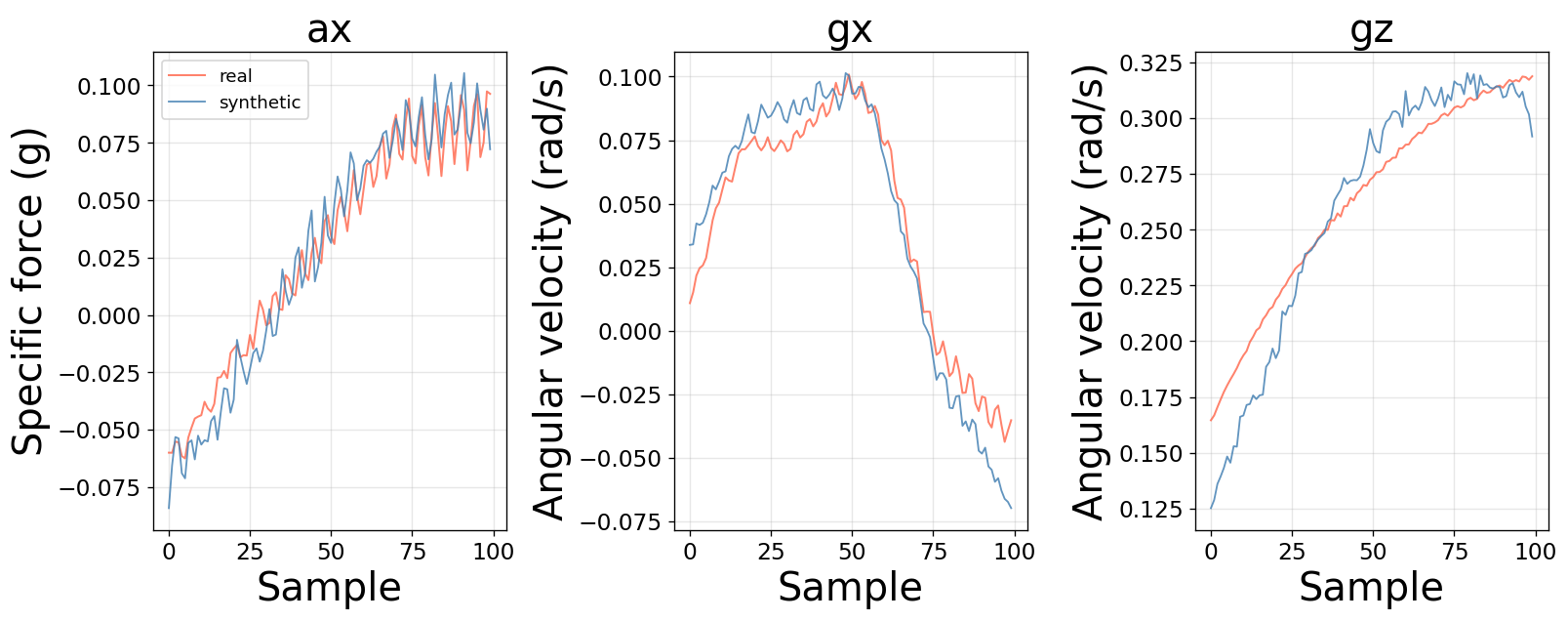}
    \caption{Pipeline 2, held-out conditioning transfer. The conditioning
statistics were taken from trajectory T13, withheld from training. Two
representative channels of the six are shown, the $x$ accelerometer
component (\text{ax}) and the $x$ gyroscope component (\text{gx}),
together with the $z$ gyroscope component (\text{gz}), a channel whose
real signal contains no local extrema and for which the peak map is empty.}
    \label{fig:p2_transfer}
\end{figure}

\begin{figure} 
    \centering
    \includegraphics[width=\linewidth]{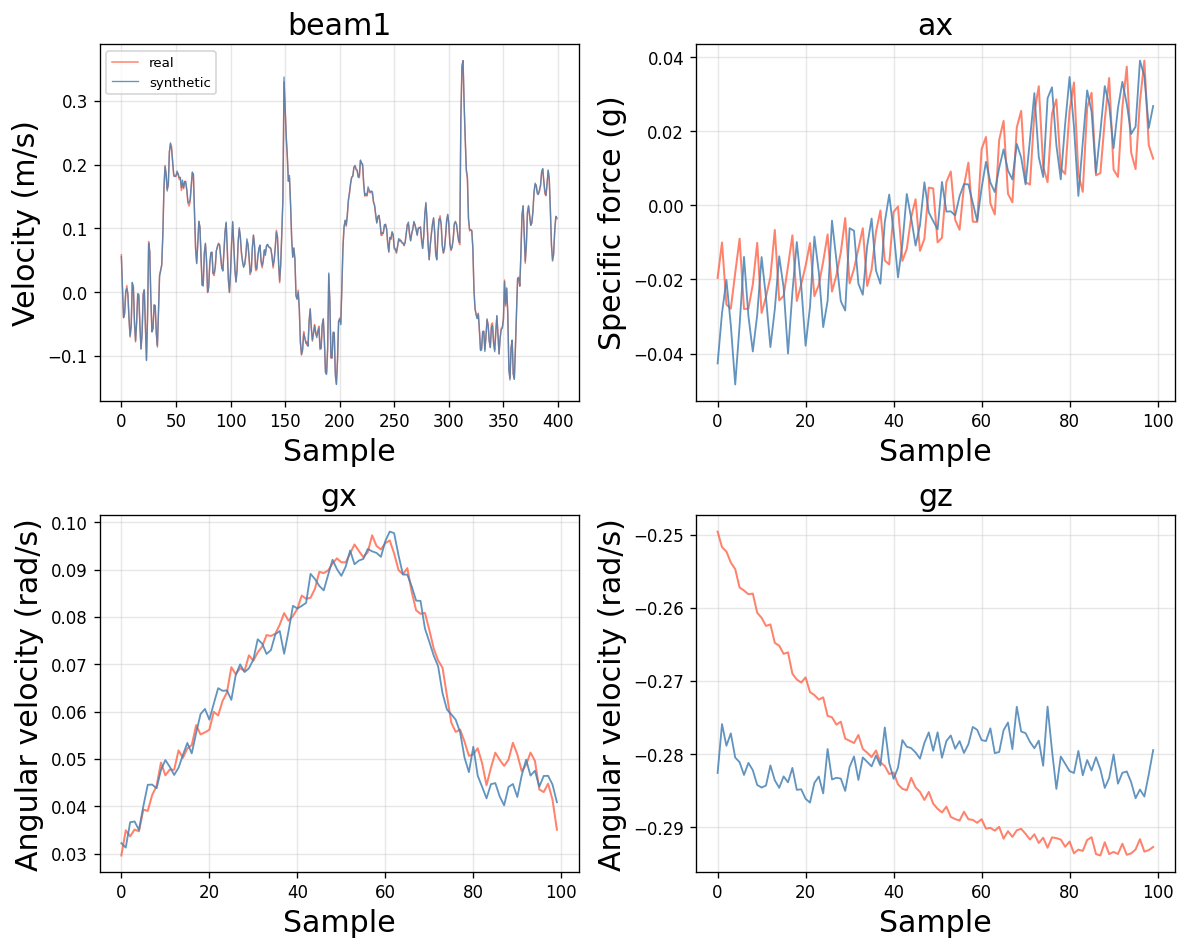}
    \caption{Pipeline 3, held-out conditioning transfer. The conditioning
statistics were taken from trajectory T5, withheld from training. Four
representative channels of the ten are shown: a DVL beam
velocity (\text{beam1}), the $x$ accelerometer component
(\text{ax}), and the $x$ and $z$ gyroscope components (\text{gx} and
\text{gz}). The $z$ gyroscope shows the largest level divergence
observed across the six inertial channels.}
    \label{fig:p3_transfer}
\end{figure}

The comparison is between a generated signal and the real signal its conditions were generated from, so it measures
condition transfer and not novelty. The conditioning statistics were taken from test trajectories, which were withheld from generator training, and the
generated signals are compared against the real recordings of those trajectories. The examples are drawn from three different test trajectories,
one per Pipeline.

In Pipeline 1 (Fig.~\ref{fig:p1_transfer}) the four beam channels are shown over a single 206-sample window. The generated signal follows the real signal
on all four beams. The transitions in vehicle motion near samples are reproduced at the correct positions on every channel, proving that the peak map is a compatible condition. Isolated sharp excursions are reproduced at the correct position but at a slightly reduced amplitude, and the generated signal departs
from the real one over the final forty samples of beam 1 and beam 4.

In Pipeline 2 (Fig.~\ref{fig:p2_transfer}) inertial channels are
generated. Three are shown: the $x$ accelerometer component
(\text{ax}) and the $x$ and $z$ gyroscope components (\text{gx} and
\text{gz}). On \text{ax} the generated signal is centered on the
conditioned mean and varies about it with the conditioned standard
deviation, and the oscillation that carries most of that variation
appears at the correct amplitude. On \text{gx} the peak is placed at
the correct position, which is the component the peak map supplies. The
two agree in window mean without agreeing point by point. On
\text{gz} the real signal is a slow monotonic rise with no local
extrema, so the peak map carries no events, and the generated signal
reproduces the rise from a lower starting value.

Pipeline 3 (Fig.~\ref{fig:p3_transfer}) generates the beam and inertial
channels jointly from a single model. Four channels are shown: the first
DVL beam velocity (\text{beam1}), the $x$ accelerometer component
(\text{ax}), and the $x$ and $z$ gyroscope components (\text{gx} and
\text{gz}). The generated signals follow the real ones on
\text{beam1}, \text{ax}, and \text{gx}, at the correct level and
amplitude, from conditioning statistics taken outside the training set.
Generating ten channels jointly therefore does not degrade reproduction
on those channels.

The \text{gz} subplot of Fig.~\ref{fig:p3_transfer} is the one case
where the generated signal departs from the real sample path. The real signal descends smoothly, while the generated signal instead varies about a value near its conditioned mean and does not descend. A monotonic descent has no local extrema, so the peak map found no events to mark. The mean fixes the average over the window, which both signals satisfy, and it does not distinguish a signal that descends through that
average from one that varies about it. The generated signal therefore satisfies the conditioning without following the real path.

The $z$ gyroscope component (\text{gz}) subplot of Fig.~\ref{fig:p2_transfer} bounds this result. That signal is also a monotonic change with no local extrema, and it is reproduced. A monotonic change is therefore not specifiable in principle under the present conditioning. 

These cases illustrate the freedom the conditioning leaves to the model. The descriptors were chosen to fix the dynamic regime of a window while leaving its
fine structure to the generator, as established in
Section~\ref{sec:preliminaries}. Each case shows one dimension that remains open under that choice. The mean fixes the average over a window and not where
within the window the signal sits, which is visible on $x$ gyroscope component (\text{gx}) in Fig.~\ref{fig:p2_transfer}. The mean and the standard deviation together fix the level and the spread and not whether the variation is a slow change or a
fast one, which is visible on $z$ gyroscope component (\text{gz}) in Fig.~\ref{fig:p3_transfer}. In generation
this is the intended behavior, since a window that satisfies the descriptors without reproducing the source signal creates a new, desired, sample. In this test the objective is reconstruction, so the same freedom appears as a departure from
the real path.

\subsubsection{Draw-to-Draw Diversity}

\begin{figure} 
    \centering
    \includegraphics[width=\linewidth]{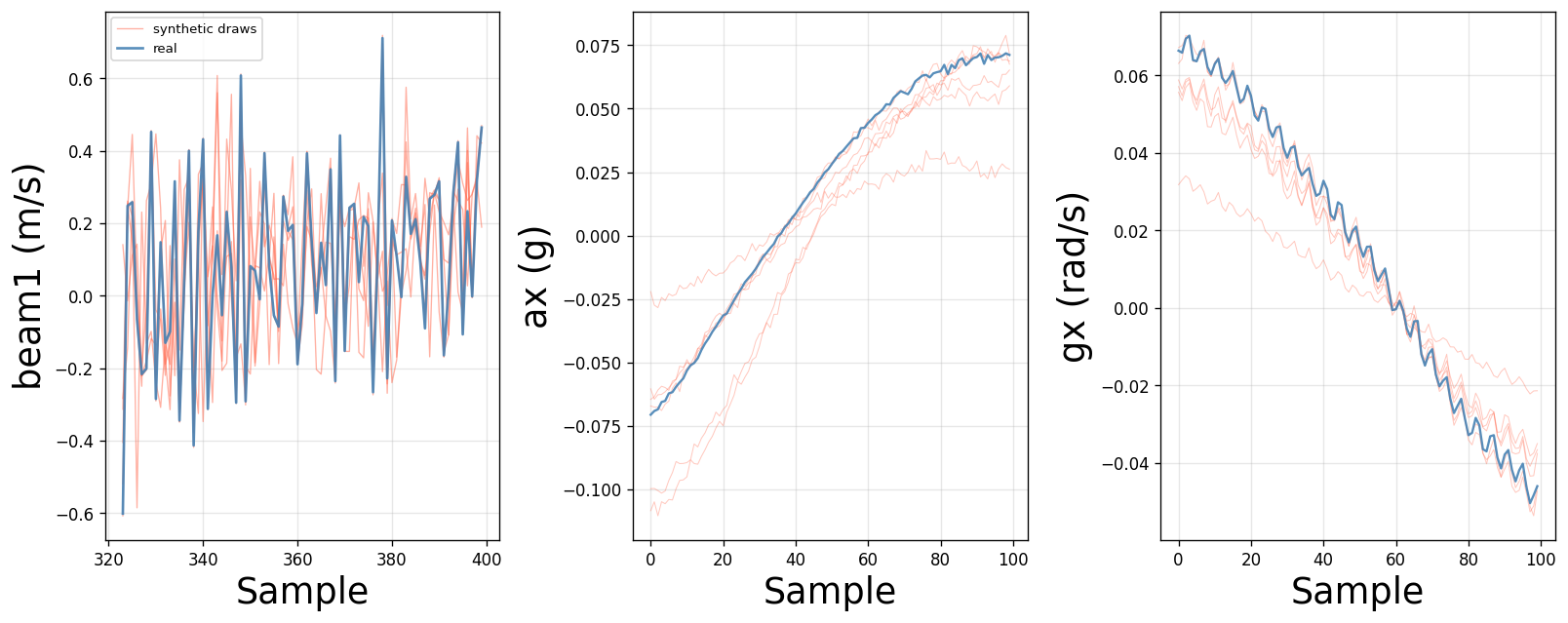}
    \caption{Example of draw-to-draw diversity in Pipeline 3, under fixed
conditioning. Two channels are shown, the $x$ accelerometer component
(\text{ax}) and the $x$ gyroscope component (\text{gx}). Several
independent samples were generated from a single conditioning vector and
are shown together with the real window that vector was computed from.}
    \label{fig:p3_diversity}
\end{figure}

Figure~\ref{fig:p3_diversity} shows an example of the signals obtained when the conditioning is fixed and the generation is repeated. Each generation begins from an
independent draw of random noise, which the denoising process transforms into a signal consistent with the requested conditions. Running the process several
times with the same conditioning vector and different starting noise therefore produces several signals that all satisfy the same requested statistics.

The draws differ for two reasons. The conditioning specifies only the mean, the
standard deviation and the peak map, so everything else about the window is
left to the model, and two draws can satisfy the same descriptors while
differing in fine structure. Separately, the generated window's actual mean and STD are not equal to the conditions, which is shown also in Section ~\ref{sec:adherence}. Two draws from one condition
can therefore differ.

\subsubsection{Nearest-Neighbour Replication}

\begin{figure*}[t]
    \centering
    \begin{subfigure}[b]{0.30\linewidth}
        \centering
        \includegraphics[width=\linewidth]{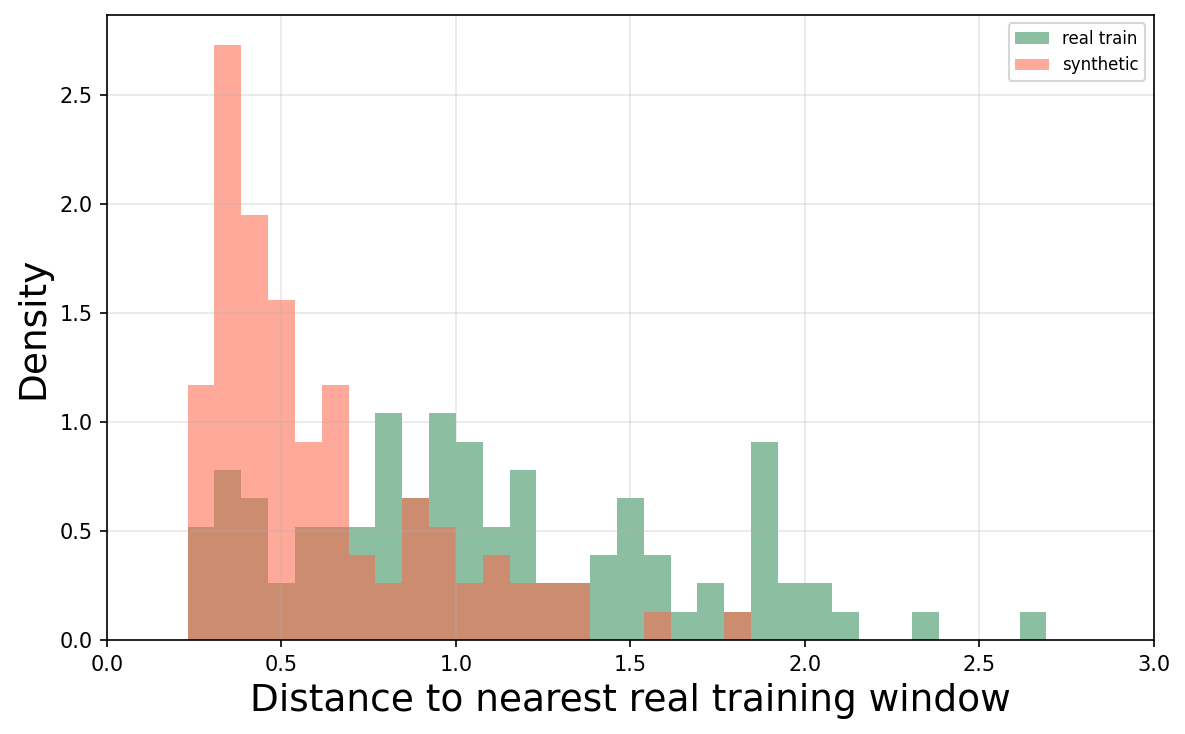}
        \caption{Pipeline 1}
        \label{fig:sub_nn_p1}
    \end{subfigure}%
    \hspace{0.02\linewidth}%
    \begin{subfigure}[b]{0.30\linewidth}
        \centering
        \includegraphics[width=\linewidth]{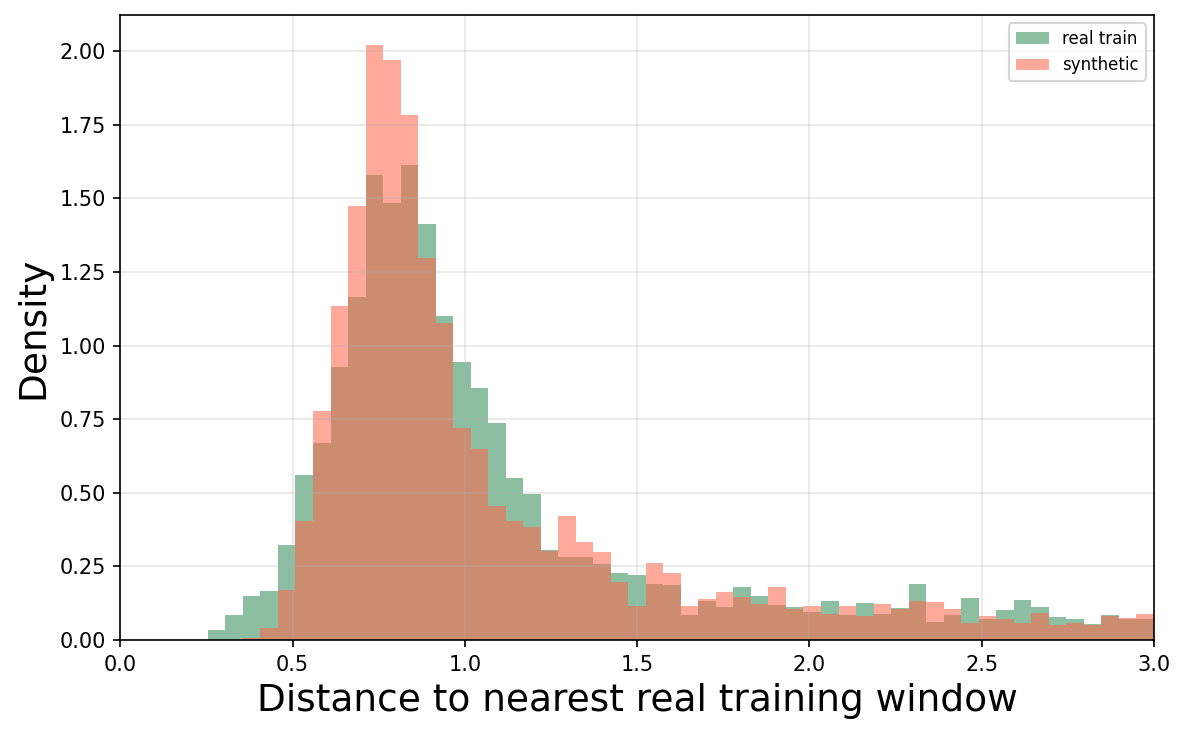}
        \caption{Pipeline 2}
        \label{fig:sub_nn_p2}
    \end{subfigure}%
    \hspace{0.02\linewidth}%
    \begin{subfigure}[b]{0.30\linewidth}
        \centering
        \includegraphics[width=\linewidth]{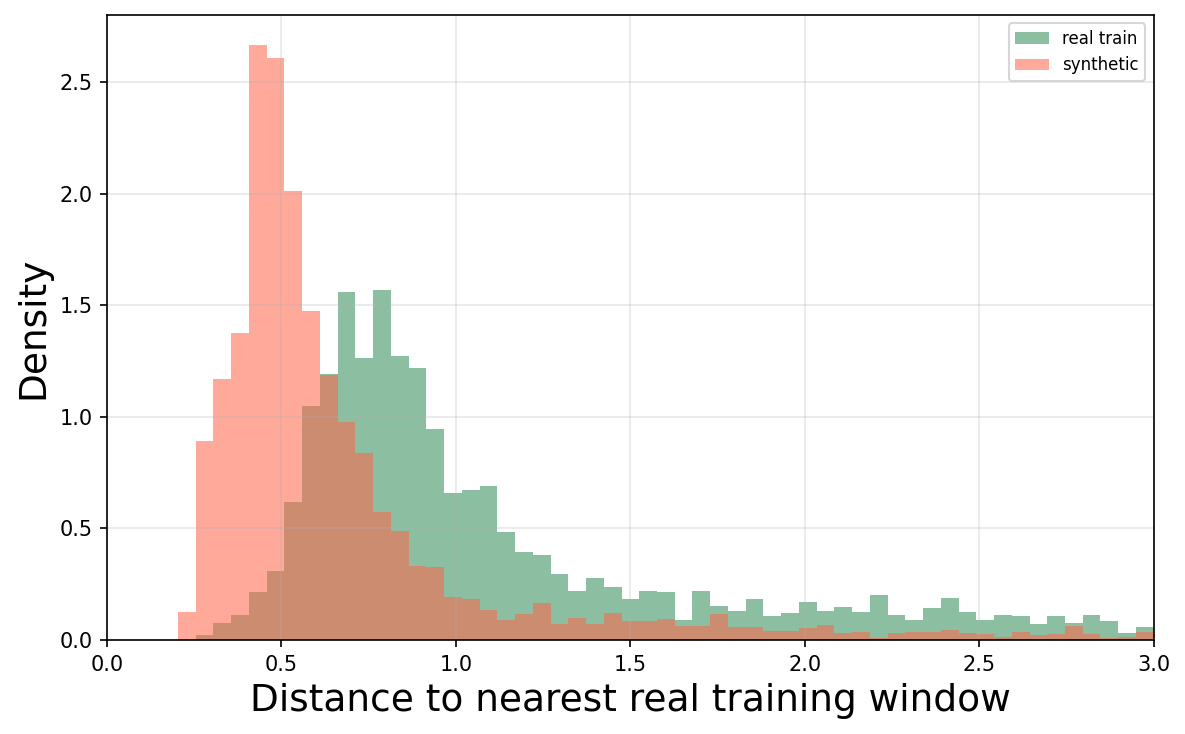}
        \caption{Pipeline 3}
        \label{fig:sub_nn_p3}
    \end{subfigure}
    \caption{Nearest-neighbour replication test. For each window, the L2
    distance to the closest window in the real training set is computed on
    per-channel windows. Green: real training windows measured against all
    other training windows, excluding themselves. This gives the natural
    nearest-neighbour scale of real data, since no real window is a copy of
    another, and it is the reference against which the synthetic distances are
    read. Red: synthetic windows measured against the real training set.}
    \label{fig:nn_replication}
\end{figure*}

Figure~\ref{fig:nn_replication} shows two distributions per Pipeline. The reference distribution, in green, is the leave-one-out distance of each real training window to the rest of the training set. It describes how far apart recorded windows naturally lie from one another, and it is the scale against
which the synthetic distances are read. The synthetic distribution, in red, is the distance from each generated window to the nearest real training window.
A copied window would lie at zero distance from the window it copied, since that window would then be its own nearest neighbour. Neither Pipeline places mass at or near zero. The closest synthetic windows lie at distances the real training windows also reach among themselves, so no generated window reproduces a recorded one. The generated data are therefore new samples rather than copies. In Pipelines 1 and 3 the synthetic distribution is shifted toward smaller distances than the reference, and the shift is largest in Pipeline 3. Generated windows therefore lie closer to the training set than real windows lie to one another, while remaining distinct from it. Pipeline 2 is the exception, and its two distributions overlap across the range.
This correlates with the conditioning scheme of Section \ref{sec:syn_data_generation}. Pipelines 1 and 3 draw most of their windows on blended conditions, which are combinations of two source conditions and therefore concentrated more centrally than the recorded ones, as established in Section \ref{sec:adherence}. Pipeline 2 conditions on single real windows and inherits the spread of the recordings.

\subsection{Downstream Results}
\label{sec:results-downstream}

The downstream results are reported as the velocity RMSE of \eqref{eq:rmse} and improvement (\ref{eq:improvement}) on three real trajectories in the test set, as described in Section \ref{sec:downstram_eval}. The reference in every experiment is the real-only baseline, which is the same network trained on the real training set alone. The architecture, the training settings and the test set are held fixed throughout, so any change in RMSE is attributable to the training data alone. 

Table~\ref{tab:downstream-summary} summarizes the downstream evaluation. Three training
set compositions are reported per Pipeline. The first is the real-only baseline. The
second adds generated windows to the full real training set at a one-to-one ratio. The
third trains on generated windows alone at five times the size of the real training set,
with no real window retained.

The generated data stands on its own. Every Pipeline reaches its baseline, and none falls below it. The margins in Pipelines 2 and 3 are small because the baseline is already near the accuracy the task permits.  Generated signals therefore substitute for real ones in the training of a downstream velocity estimator. This holds across three Pipelines, two BeamsNet architectures, and both sensors, up to the complete removal of the recorded training data.

\begin{table*}[t]
\centering
\caption{Downstream velocity RMSE on the held-out real trajectories for the three Pipelines, under three training set compositions. Improvement is relative to the real-only baseline of the same Pipeline. Pipeline 2
has no fully synthetic entry and cannot exceed a 1:1 ratio. It generates
the inertial channels alone, and each synthetic inertial window is
paired with a real DVL window, so every training window contains real
beams and no more synthetic windows can be produced than there are real
DVL windows.}
\label{tab:downstream-summary}
\begin{tabular}{|l|c|c|c|c|c|c|}
\hline
\multirow{2}{*}{\textbf{Training set}} & \multicolumn{2}{c|}{\textbf{Pipeline 1}} & \multicolumn{2}{c|}{\textbf{Pipeline 2}} & \multicolumn{2}{c|}{\textbf{Pipeline 3}} \\
\cline{2-7}
 & RMSE [m/s] & Improv. & RMSE [m/s] & Improv. & RMSE [m/s] & Improv. \\
\hline
Real only (baseline)    & 0.0039 & --       & 0.0036 & --      & 0.0036 & --      \\
\hline
Real + synthetic 1:1 & 0.0036 & 7.7\,\%  & 0.0035 & 2.5\,\% & 0.0036 & 1.7\,\% \\
\hline
Synthetic only 5:1   & 0.0035 & 10.3\,\% & --     & --      & 0.0036 & 1.5\,\% \\
\hline
\end{tabular}
\end{table*}
\subsubsection{Fully Synthetic Training}
\label{sec:results-fullysynth}

\begin{figure*}[t]
    \centering
    \begin{subfigure}[t]{0.4\linewidth}
        \centering
        \includegraphics[width=\linewidth]{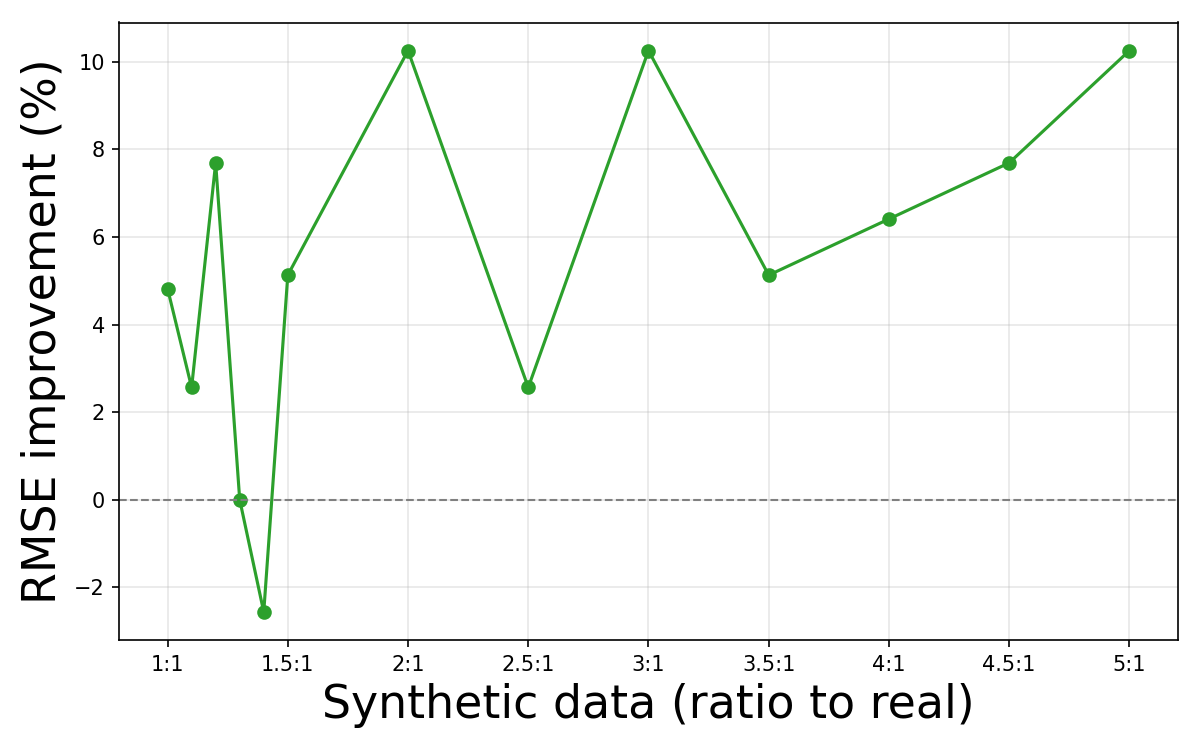}
        \caption{Pipeline 1}
        \label{fig:fully_syn_p1}
    \end{subfigure}
    \hfill
    \begin{subfigure}[t]{0.4\linewidth}
        \centering
        \includegraphics[width=\linewidth]{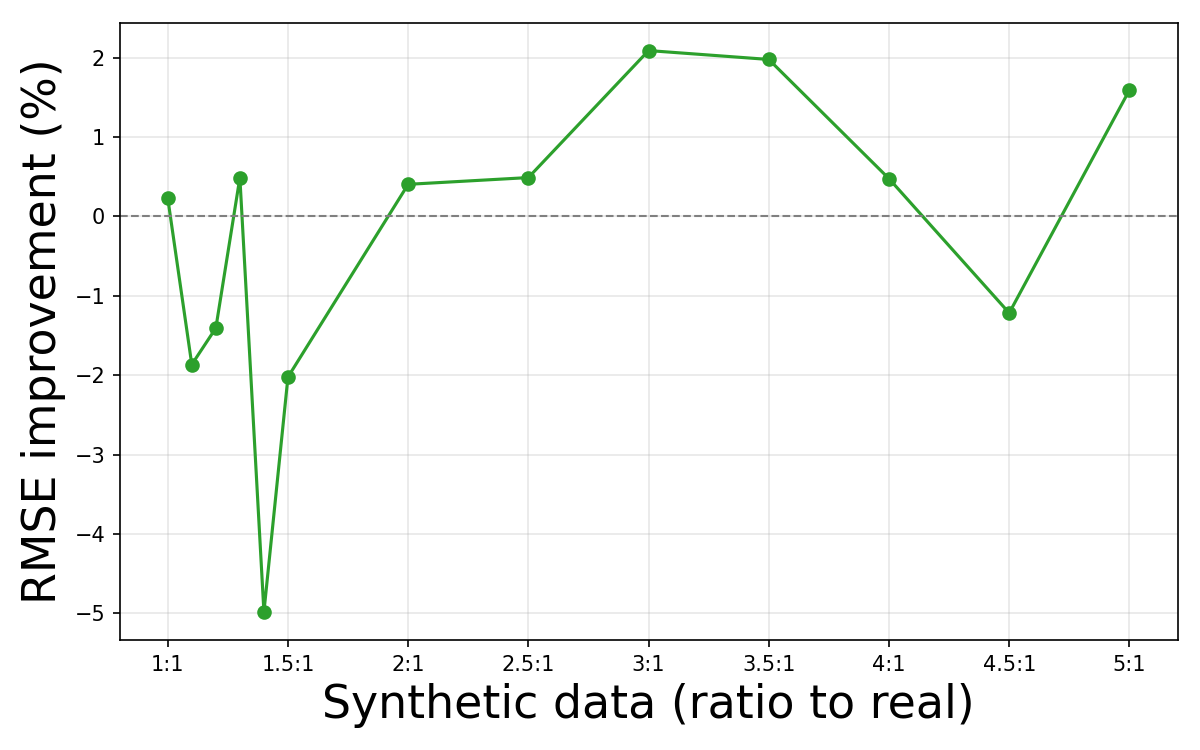}
        \caption{Pipeline 3}
        \label{fig:fully_syn_p3}
    \end{subfigure}
    \caption{RMSE improvement of a downstream model trained entirely on synthetic data, as a function of synthetic data size, relative to a real-only baseline.}
    \label{fig:fully_synthetic}
\end{figure*}

Figure~\ref{fig:fully_synthetic} reports the RMSE of a network trained on synthetic windows alone, as the synthetic dataset is swept from 1:1 to 5:1 relative to the size of the real training dataset. Pipeline 2 is excluded, since every training window in that Pipeline contains real beams by construction. The reference is the accuracy of the real-only baseline.

Pipeline 3 is at the baseline already at 1:1, and the curve stays within a few percent of it across the swept range with no monotonic trend. Pipeline 1 is likewise at the baseline at 1:1 and holds that level to 5:1. The excursions above the baseline in either Pipeline are of the same size as the point-to-point scatter along the curves, so they are not read as an improvement
over training on real recordings. The result is that generated beams alone are sufficient to train BeamsNetV2, and ten jointly generated channels are sufficient to train BeamsNetV1.

\subsubsection{Combined Synthetic and Real Training}
\label{sec:results-additive}

\begin{figure*}[t]
    \centering
    \begin{subfigure}[b]{0.30\linewidth}
        \centering
        \includegraphics[width=\linewidth]{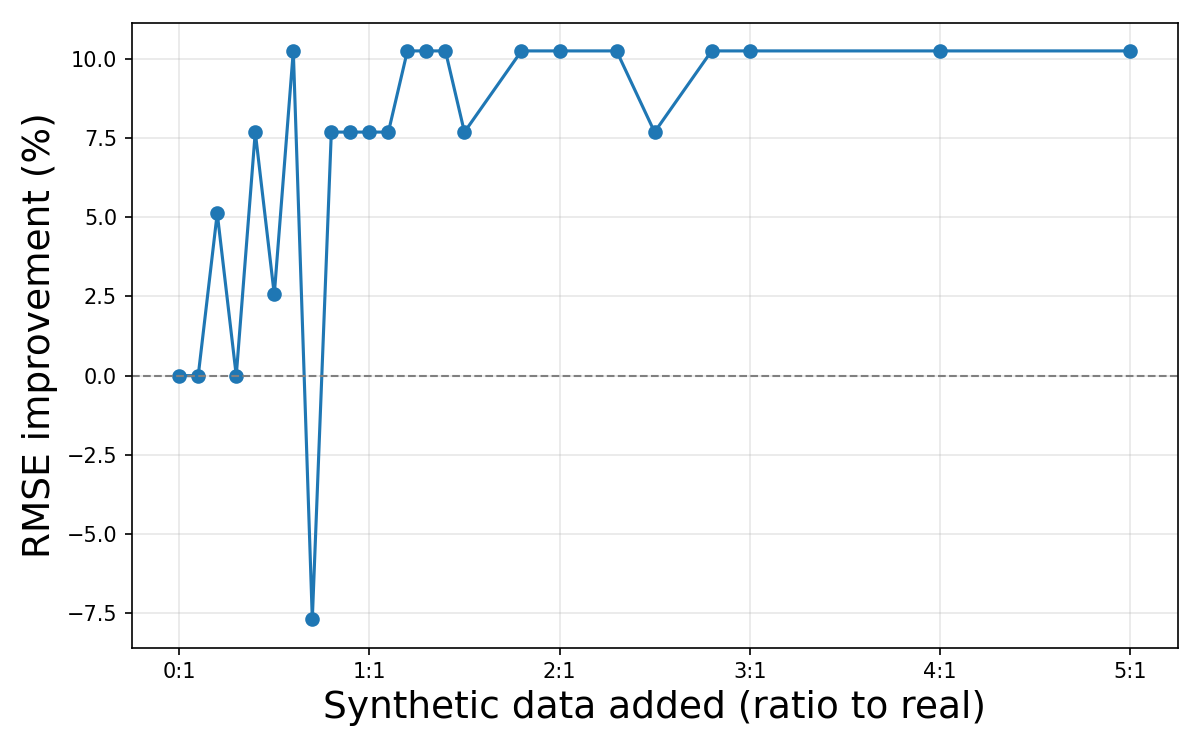}
        \caption{Pipeline 1}
        \label{fig:additive_p1}
    \end{subfigure}%
    \hspace{0.02\linewidth}%
    \begin{subfigure}[b]{0.30\linewidth}
        \centering
        \includegraphics[width=\linewidth]{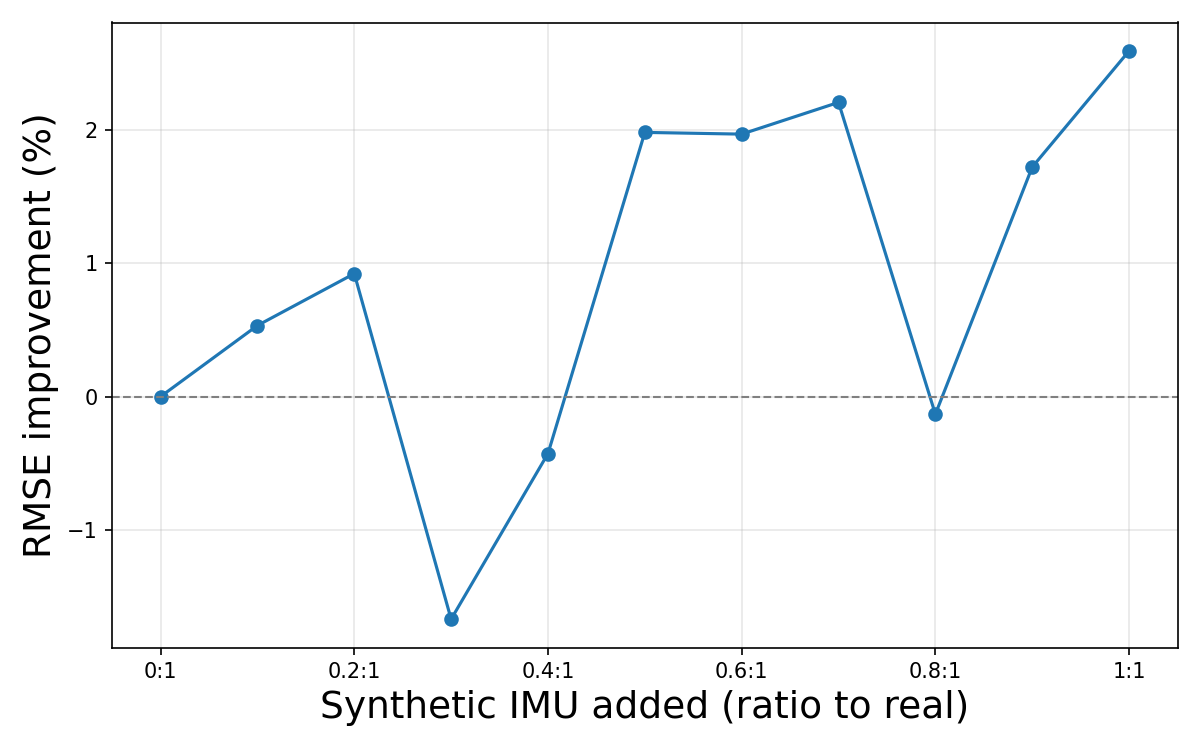}
        \caption{Pipeline 2}
        \label{fig:additive_p2}
    \end{subfigure}%
    \hspace{0.02\linewidth}%
    \begin{subfigure}[b]{0.30\linewidth}
        \centering
        \includegraphics[width=\linewidth]{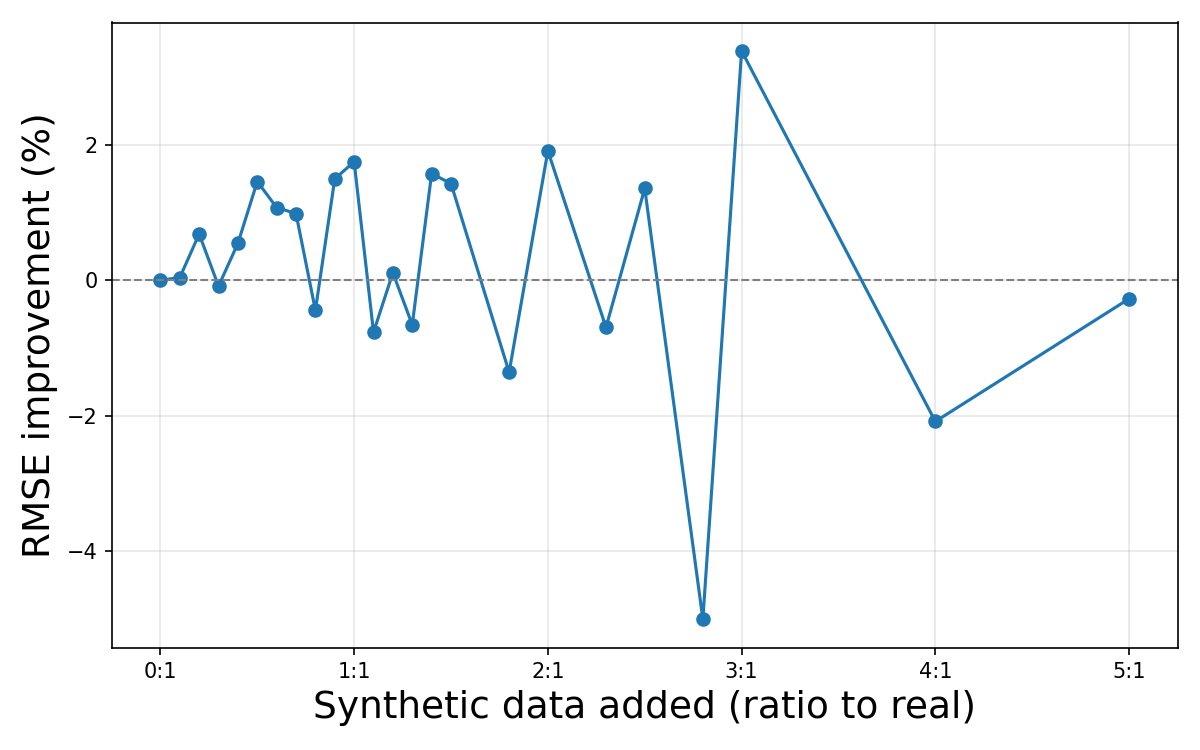}
        \caption{Pipeline 3}
        \label{fig:additive_p3}
    \end{subfigure}
    \caption{RMSE improvement when synthetic data is added to the full real training data, as a function of the amount of synthetic data added.}
    \label{fig:additive}
\end{figure*}

Figure~\ref{fig:additive} reports the RMSE when generated windows are added to the full
real training set, as the amount added is swept from 0:1 to 5:1 relative to its size.
Pipeline 2 is swept to 1:1 only, since its synthetic pool is bounded by the number of
real DVL windows it is paired with.

The Pipelines differ in how much they respond. In Pipeline 1 the RMSE falls by about ten percent relative to the baseline and holds that level from 1.15:1 onward. Pipelines 2 and 3 remain within a few percent of the baseline across the swept range, with no monotonic dependence on the amount added. The result is that generated windows supplement the real training set in Pipeline 1, and that they neither improve nor degrade BeamsNetV1 in Pipelines 2 and 3.

\subsubsection{Real-to-Synthetic Substitution}
\label{sec:results-substitution}

\begin{figure*}[t]
    \centering
    \begin{subfigure}[b]{0.30\linewidth}
        \centering
        \includegraphics[width=\linewidth]{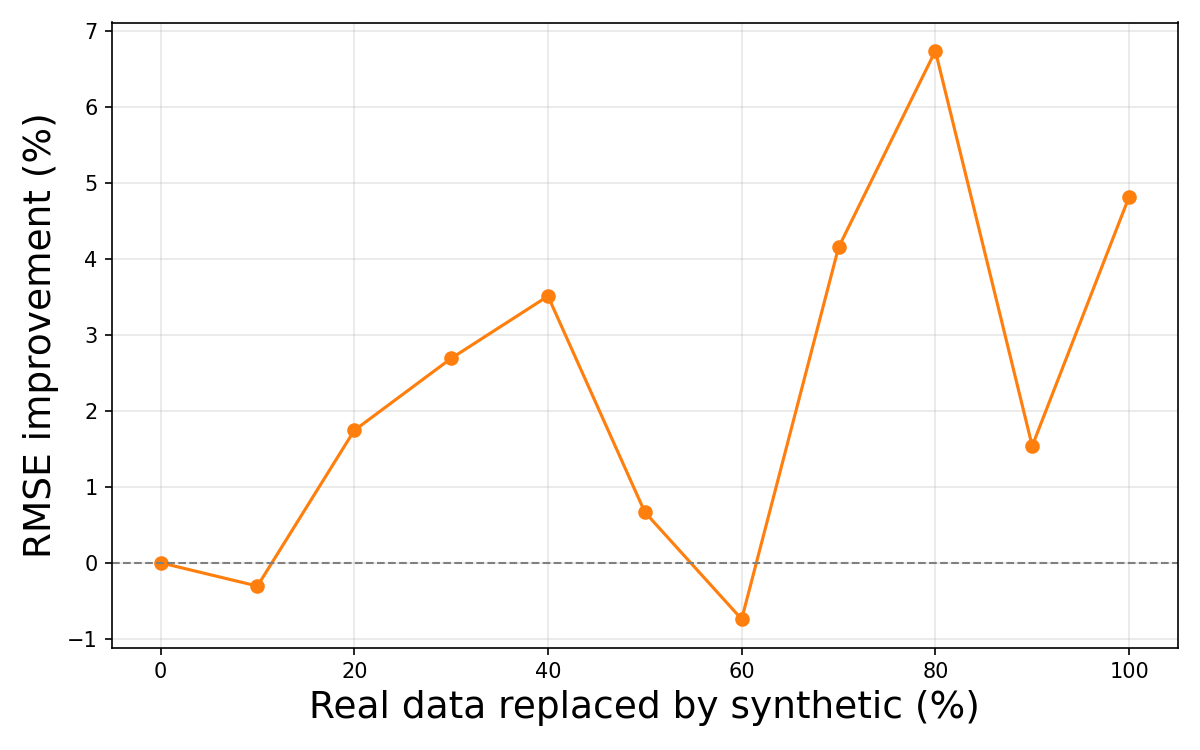}
        \caption{Pipeline 1}
        \label{fig:sub_p1}
    \end{subfigure}%
    \hspace{0.02\linewidth}%
    \begin{subfigure}[b]{0.30\linewidth}
        \centering
        \includegraphics[width=\linewidth]{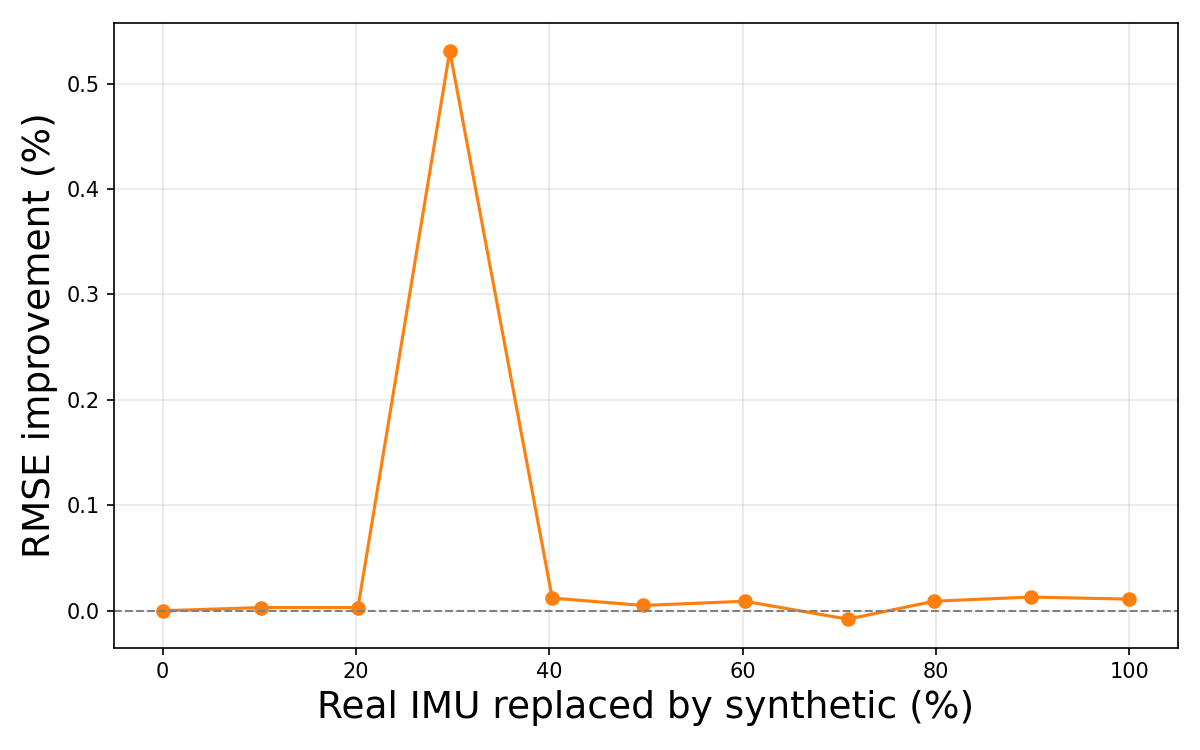}
        \caption{Pipeline 2}
        \label{fig:sub_p2}
    \end{subfigure}%
    \hspace{0.02\linewidth}%
    \begin{subfigure}[b]{0.30\linewidth}
        \centering
        \includegraphics[width=\linewidth]{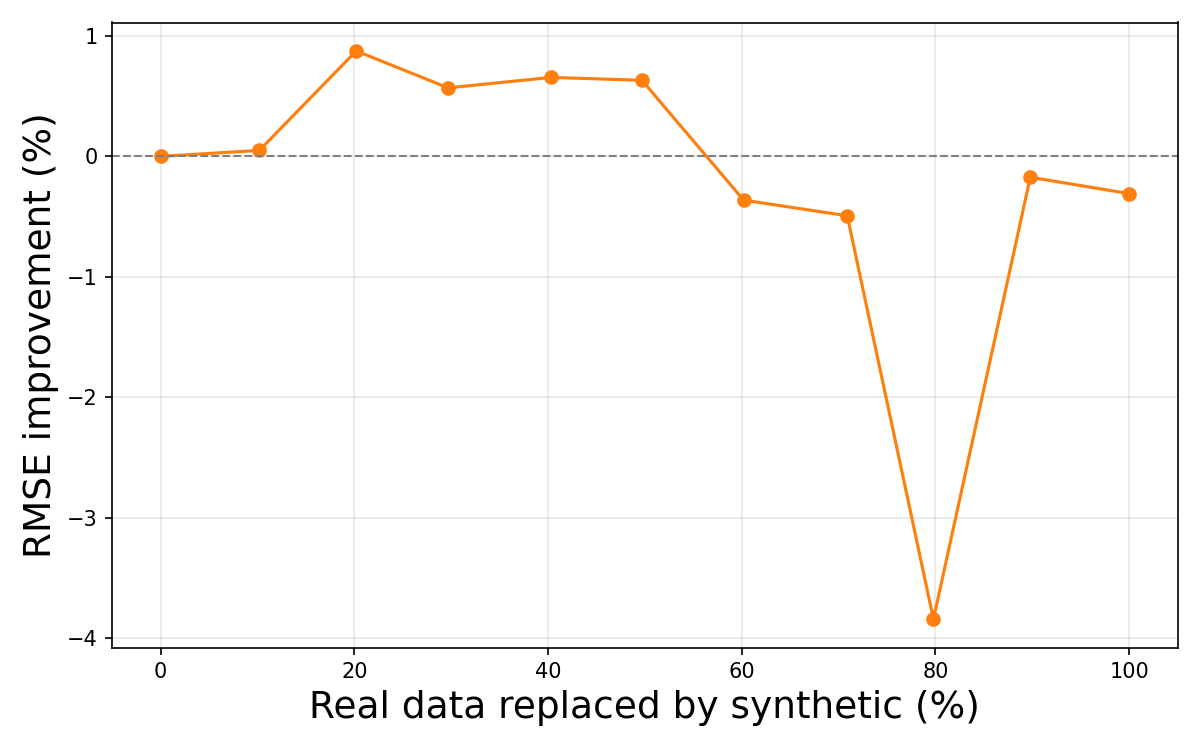}
        \caption{Pipeline 3}
        \label{fig:sub_p3}
    \end{subfigure}
    \caption{RMSE improvement when a fraction of the real training data is replaced by an equal amount of synthetic data, from $0\%$ (real-only baseline) to $100\%$ (fully synthetic, same size as the real-only dataset).}
    \label{fig:real_to_synthetic}
\end{figure*}

Figure~\ref{fig:real_to_synthetic} holds the total number of training windows fixed and replaces a fraction of the real windows with an equal number of generated windows, from $0\%$ to $100\%$. 

In all three Pipelines the RMSE remains within a few percent of the baseline at every substitution fraction, including full replacement. No Pipeline shows a degradation that grows with the substitution fraction. A single point in Pipeline 3, falls about $4\%$ below the baseline, and the curve returns to the baseline at $100\%$. These figures show that a synthetic signal can be exchanged for a real signal in the BeamsNet training set without a measurable loss of accuracy on the held-out real trajectories, up to complete replacement of the real training data.

\subsubsection{Conditioning Ablation}
\label{sec:results-ablation}

Table~\ref{tab:ablation} reports the conditioning ablation, as established in Section \ref{sec:downstram_eval}. In Pipelines 1 and 3, the mean and STD conditioning carry the main weight of creating plausible signals. Removing it raises the RMSE by more than two orders of magnitude in Pipeline 1 and by a factor of about fifty in Pipeline 3, and removing all conditioning has the same effect. 
The peak map locates the variation in time and does not set the level of the window, so its removal degrades the generated windows without breaking their correspondence to the reference velocity they are paired with. The difference between the Pipelines is correlated with the window construction: the DVL beam windows of Pipeline 1 span several maneuvers, while the joint IMU and DVL windows of Pipeline 3 cover a single DVL epoch, meaning that less temporal structure is available
for the descriptor to locate. Its contribution in Pipeline 1 is visible in the held-out conditioning transfer of Section~\ref{sec:results-transfer}, where the transitions in vehicle motion are reproduced at the correct positions on every beam channel.
In Pipeline 2, the ablation leaves the downstream RMSE unchanged. The DVL beams supplied to BeamsNetV1 are real in this Pipeline, and the generated inertial channels are a secondary input to the velocity estimate, so degrading the inertial generator does not propagate
into a measurable change in RMSE. The downstream RMSE is therefore not a sensitive test of the conditioning in Pipeline 2.

\begin{table*}[t]
\centering
\caption{Conditioning ablation across the three Pipelines. Downstream velocity RMSE for the corresponding downstream model trained on synthetic data at a one-to-one ratio to real data, for each conditioning variant. Change is relative to the full conditioning of the same Pipeline.}
\label{tab:ablation}
\small
\begin{tabular}{|l|c|c|c|c|c|c|}
\hline
\multirow{2}{*}{\textbf{Conditioning}} & \multicolumn{2}{c|}{\textbf{Pipeline 1}} & \multicolumn{2}{c|}{\textbf{Pipeline 2}} & \multicolumn{2}{c|}{\textbf{Pipeline 3}} \\
\cline{2-7}
 & \textbf{RMSE [m/s]} & \textbf{Change} & \textbf{RMSE [m/s]} & \textbf{Change} & \textbf{RMSE [m/s]} & \textbf{Change} \\

\hline
Full conditioning & 0.003712 & 0\%           & 0.003525 & 0\%       & 0.003608 & 0\% \\
\hline
No conditioning   & 0.9622   & $-25{,}822\%$ & 0.003526 & $-0.03\%$ & 0.262380 & $-7{,}171\%$ \\
\hline
No peak map       & 0.0043   & $-15.8\%$     & 0.003526 & $-0.03\%$ & 0.003626 & $-0.50\%$ \\
\hline
No mean and STD   & 1.009    & $-27{,}079\%$ & 0.003526 & $-0.03\%$ & 0.178263 & $-4{,}840\%$ \\
\hline

\hline
\end{tabular}
\end{table*}

\section{Conclusion}
\label{sec:conclusion}
This paper presents a conditional diffusion approach for generating DVL and inertial signals for AUV navigation. To this end, synthetic signals were created by a conditional diffusion model trained on real recordings, under an EDM formulation, and evaluates them in three configurations that differ in which sensor is generated. In Pipeline 1 the beam velocities alone are generated and used to train BeamsNetV2. In Pipeline 2 the inertial channels alone are generated and paired with real DVL beams to train BeamsNetV1. In Pipeline 3 both DVL beams and inertial sensors are generated jointly by a single model and used to train BeamsNetV1.

The conditioning is the novel design element that makes the diffusional generation controllable. Three descriptors are supplied per channel: the mean and the standard deviation, which fix the amplitude of the motion, and a peak map, which marks the position, sign, and strength of maneuver events within the window. The descriptors were purposefully chosen to be low-dimensional. The scalar descriptors and the peak map enter the network by different mechanisms, since they differ in form: the mean and STD modulate every residual block through FiLM, while the peak map is a time-varying channel concatenated at the input and injected at each encoder stage. In the joint Pipeline, the DVL beam channels and the inertial channels are given separate conditioning pathways, so that the scale and shift applied to a beam channel are computed from the beam conditions alone.

The generated signals were assessed at the signal level by four tests. The conditioning input reaches the output in all three Pipelines, with no channel approaching the zero correlation that would indicate the condition was ignored. The same test bounds where the conditioning is reliable: adherence is accurate in the densely populated region of the condition distribution and regresses toward the marginal mean at the edges of the range. The effect is strongest on the gyroscope channels, whose requested values concentrate near zero, and weakest in Pipeline 2, which has the largest training set relative to the number of channels it generates. The conditioning mechanism transfers to trajectories withheld from generator training, proving it is not fitted to the conditions present in the training recordings. Repeated draws from one condition are distinct on every channel, proving the descriptors leave the generator the freedom they were designed to leave. No generated window reproduces a training window, which establishes that the synthetic data are new samples rather than copies.

At the downstream task level, tested on the BeamsNet models, generated signals substitute for real ones in the training. Both fully synthetic Pipelines reach the real-only baseline at a 1:1 ratio to the real training pool and hold that level to 5:1, with no real window retained. Substitution at a fixed training set size costs no measurable accuracy at any replacement fraction, including complete replacement. The points above the baseline are of the same size as the point-to-point scatter along the curves, meaning the result is stated as a match to the real dataset. This work shows that a downstream velocity estimator trained without a single real recording reaches the accuracy of the same estimator trained on the full real training set, and it does so across three Pipelines, two downstream architectures, and both sensors. 
The conditioning ablation confirms that all three descriptors are operative and separates their contributions. The mean and standard deviation carry the main weight in Pipelines 1 and 3. Removing them raises the RMSE by more than two orders of magnitude in Pipeline 1 and by a factor of about fifty in Pipeline 3, and removing all conditioning has the same effect, so the generated signals depend on the conditioning rather than on the marginal distribution the model learned. 

These results are bounded by the evaluation setting. The A-KIT dataset provides thirteen trajectories, of which ten are available for training, and the downstream evaluation rests on two architectures from a single work. Taken together, the two levels of evaluation support the following conclusion: the signal-level tests establish that the conditioning is operative, that it transfers to trajectories the generator never saw, and that the generated windows are distinct from one another and from the recordings. The downstream tests establish that windows with these properties train a velocity estimator to the accuracy obtained from real recordings. A small set of motion descriptors is therefore sufficient to steer a diffusion model toward synthetic DVL and inertial
signals that carry the information the downstream task requires, and those signals stand in for recorded ones up to the complete removal of the real training data.

Future work follows from these bounds. Larger recording sets and additional sensor types would test the conditioning design beyond the present dataset. A downstream task in which the inertial channels carry more weight than they do in BeamsNetV1 would test the generated inertial signals where the present benchmark cannot. 

Overall, by reducing the reliance of learned AUV navigation on sea-trial data and enabling training under scarce-recording conditions, the proposed approach paves the way for more accessible development and evaluation of deep learning methods in underwater navigation.

\printbibliography

\end{document}